\documentclass{article}
\usepackage[margin=0.5in]{geometry}

\title{Short-term Traffic Prediction with Deep Neural Networks: A Survey}
\author{Kyungeun Lee$^{1}$, Moonjung Eo$^{1}$, 
    Euna Jung$^{1}$, Yoonjin Yoon$^{2}$,
    and Wonjong Rhee$^{1, \ast}$ \\
    \small $^{1}$Department of Intelligence and Information, Seoul National University, Seoul, Korea \\
    \small $^{2}$Department of Civil and Environmental Engineering, KAIST, Daejeon, Korea\\
    \small $^\ast$Corresponding author: Wonjong Rhee, wrhee@snu.ac.kr}
\date{}

\usepackage{graphicx}
\usepackage{adjustbox}
\usepackage{amssymb}
\usepackage{lineno}
\usepackage{mathtools}
\usepackage{amsmath}
\usepackage{amsfonts}
\usepackage{subfig}
\usepackage{multirow}
\usepackage{cite}

\usepackage{longtable}
\usepackage{wrapfig}
\usepackage{sidecap}
\usepackage{array}
\usepackage{url}  
\usepackage{booktabs}
\usepackage{amsthm}

\newcommand{\midsepremove}{\aboverulesep=0mm \belowrulesep=0mm}
\newcommand{\midsepdefault}{\aboverulesep=0.605mm \belowrulesep=0.984mm}

\providecommand{\keywords}[1]
{
  \small	
  \textbf{\textit{Keywords---}} #1
}

\begin{document}

\maketitle

\begin{abstract}
In modern transportation systems, an enormous amount of traffic data is generated every day. This has led to rapid progress in short-term traffic prediction (STTP), in which deep learning methods have recently been applied. In traffic networks with complex spatiotemporal relationships, deep neural networks (DNNs) often perform well because they are capable of automatically extracting the most important features and patterns. 
In this study, we survey recent STTP studies applying deep networks from four perspectives. 1) We summarize input data representation methods according to the number and type of spatial and temporal dependencies involved. 2) We briefly explain a wide range of DNN techniques from the earliest networks, including Restricted Boltzmann Machines, to the most recent, including graph-based and meta-learning networks. 3) We summarize previous STTP studies in terms of the type of DNN techniques, application area, dataset and code availability, and the type of the represented spatiotemporal dependencies. 4) We compile public traffic datasets that are popular and can be used as the standard benchmarks. Finally, we suggest challenging issues and possible future research directions in STTP.
\end{abstract}


\section{Introduction}
\label{sec:intro}

Advances in transportation systems have resulted in the generation of a large amount of traffic data from various sources \cite{veres2019survey,zhu2018itsbigdata,zhang2011itsdata}. In everyday life, GPS sensors installed in smartphones carried by millions of people can collect \textit{crowd flow} data. Furthermore, taximeters and bus card readers can collect \textit{crowd demand} data, and vehicle loop detectors can collect \textit{traffic flow} or \textit{speed} data. In the mean time, Deep Neural Networks (DNNs) have achieved promising performance improvements in various application areas. They can classify images into thousands of classes \cite{touvron2020imagenet,xie2020imagenet2} as well as recognize human speech \cite{saon2017speech,xiong2016speech2}, with only small errors. Owing to the availability of large traffic datasets and the advances in DNN techniques, DNNs have recently been applied to Short-Term Traffic Prediction (STTP).

Compared with classic machine learning algorithms, such as linear regression and Auto-Regressive Integrated Moving Averages (ARIMA), DNNs can fit a wide variety of functions. A typical DNN model has hundreds of thousands or even tens of millions of parameters that are learnt during training. Such a large number of parameters would result in overfitting in classical machine learning algorithms. By contrast, DNNs are known to avoid overfitting with a reasonable amount of data \cite{zhang2016rethinking}. Thus, based on large datasets, DNNs could learn any highly complex non-linear functions without  serious overfitting issues.

Although DNNs can automatically extract the underlying features, in practice, they are fed with appropriate inputs, the representations of which can be matched with the intended \textit{inductive bias} and deep network architecture \cite{battaglia2018inductivebias}. \textit{Inductive bias} induces DNNs to prioritize one solution over another by imposing constraints on relationships and interactions among entities in a learning process based on prior knowledge. Determining the type of inductive bias to be induced affects the DNN architecture design, and therefore it also affects the final prediction performance. For instance, in \cite{tobler1970computer}, one of the most famous priors for the spatial dependency was proposed:  \textit{``Everything is related to everything else, but near things are more related than distant things''}. If this prior knowledge is to be used as DNN inductive bias, we should investigate \textit{which input data format} could properly represent locality, and \textit{which DNN type} could effectively capture the represented locality. Intuitively, grid-based image-like representations, such as satellite photographs, may be able to represent locality, as they convert proximate regions into adjacent pixels. In addition, convolutional layers can effectively capture image locality by training adjacent pixels using a single local filter. 

In machine learning research, standard benchmark datasets are often used so that algorithms can be easily compared, and studies easily combined. For instance,  CIFAR-10\cite{cifar10dataset}/ CIFAR-100\cite{cifar100dataset} and ImageNet\cite{imagenet_cvpr09} are known as the benchmark datasets for visual tasks, and they have greatly contributed to the recent progress in this area. In STTP, however, there is \textit{no} standard benchmark dataset. 
If there were such a dataset, related research would be significantly accelerated because findings would be easy to share and integrate. 
Accordingly, we summarize public datasets that have been widely used in previous studies. In addition, we summarize papers with publicly available code.

The remainder of this paper is organized as follows. Section \ref{sec:rep} presents input data representation methods for spatial and temporal dependencies of traffic networks. Section \ref{sec:dnn} overviews deep learning techniques in chronological order. Section \ref{sec:appl} summarizes previous work according to the application area. Section \ref{sec:dataset} introduces open datasets important for generalizability. Section \ref{sec:discussion} discusses current challenges and possible research directions in STTP.

\section{Input Data Representation Methods}
\label{sec:rep}

Short-term traffic prediction is based on spatiotemporal information of a traffic network consisting of several spatial units (e.g., links, roads segments, and regions). Early studies assumed that it suffices to represent spatiotemporal information as independent features without considering any dependencies between spatial or temporal units. For example, to predict future crowd flow, \cite{liu2017crowdflow} generated independent features, such as inbound or outbound, holiday or not, weekday, and hours. 

However, a traffic network includes complex and non-linear relationships between spatial and/or temporal units. In theory, deep networks can learn any latent relationship between input units in the dataset. In practice, however, the actual performance after training is heavily dependent on the input data representation, which is therefore an important matter. In addition, some networks impose important relational inductive biases, which can be effectively provided through a specific type of input data representation \cite{battaglia2018inductivebias}. For instance, convolutional layers provide locality and translation invariance as a relative inductive bias. Thus, input data should be arranged to make local units proximate using grid representation. Otherwise, recurrent layers carry sequentiality and temporal invariance as a relative inductive bias. Thus, input data should be aligned in a temporal order. Deep network architectures are explained in more detail in Section \ref{sec:dnn}. 

Herein, we summarize the representation of spatial and temporal dependencies in STTP according to the \textit{number} and \textit{type} of dependencies that are manipulated in a sample. In general, one dependency corresponds to one data dimension. We note that external information (e.g., weather conditions, temperature, and wind speed) is not mentioned because it is considered sufficient to represent such information as independent features only in most studies.

\subsection{Representing Spatial Dependencies}
\subsubsection{Stacked Vector}

\begin{figure}[!h]
\centering
\subfloat[Example of traffic network in which it is easy to stack the road segments in clockwise connection order.]{
    \includegraphics[width=0.2\textwidth]{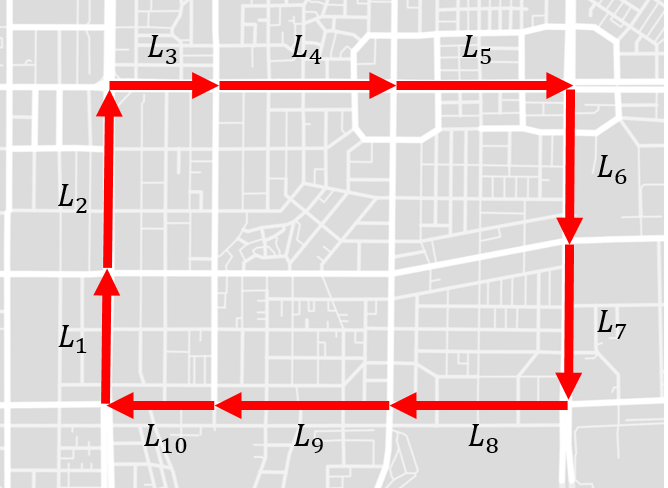}
    \label{fig:stackedvector-a}
} \hspace{0.2cm}
\subfloat[Example of traffic network in which a stacking method for input data representation is not easily applied.]{
    \includegraphics[width=0.2\textwidth]{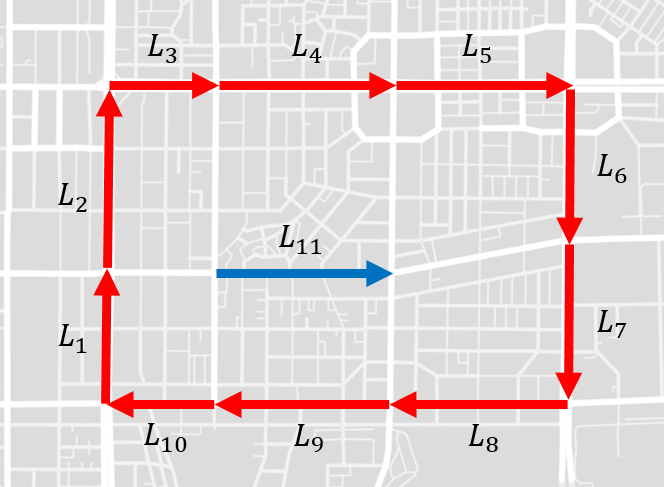}
    \label{fig:stackedvector-b}
}
\caption{Example of stacked-vector method for input data representation.}
\label{fig:stackedvector}
\end{figure}

To represent a spatial dependency, we can simply stack the data of spatial units into a single vector according to a predefined rule depending on domain knowledge or personal preference. For instance, to represent the traffic network $L=\left\{L_1,…,L_{10}\right\}$  ($L_i$ indicates a road segment), as shown in Fig. \ref{fig:stackedvector-a}, we can simply stack the road segments in clockwise connection order. Then, the final representation is $[ V(L_1 ),V(L_2 ),…,V(L_{10}) ]$, where $V(L_i)$ is the traffic value of $L_i$ at a specific time. We found some general representation rules, such as connection order, instream/outstream(inflow/outflow) order, and random listing.
This method is quite simple and useful when the network has a powerful dependency. However, it does not provide general procedures for determining the representation in general networks. As a simple example, it is not trivial to determine the dependency representation rule in the case of Fig. \ref{fig:stackedvector-b}.

\vspace{-0.2cm}

\begin{table*}[!ht]
\caption{Summary of representation methods. $N$ is the number of spatial or temporal dependencies that are considered for forming the input data representation.}
\resizebox{\textwidth}{!}{%
\begin{tabular}{ccc|ccc}
\hline
\multicolumn{3}{c|}{\textbf{Spatial Dependency}}                                                                                                                                                            & \multicolumn{3}{c}{\textbf{Temporal Dependency}}                                                                                                                               \\ \hline
N & Representation Methods                                                                                                                      & Abbr.                                                      & N & Representation Methods                                                                                                 & Abbr.                                               \\ \hline
1 & \begin{tabular}[c]{@{}c@{}}Stacked vector (random)   \\ Stacked vector (stream flow)\\ Graph (considering single dependency)\end{tabular} & \begin{tabular}[c]{@{}c@{}}1-SR\\ 1-SF\\ 1-Gr\end{tabular} & 1 & Sequentiality                                                                                                        & 1-S                                                 \\ \hline
2 & Grid representation (coordinates)                                                                                                         & 2-G                                                        & 2 & \begin{tabular}[c]{@{}c@{}}Sequentiality and periodicity  (weekly)\\ Sequentiality and periodicity (daily)\end{tabular} & \begin{tabular}[c]{@{}c@{}}2-SW\\ 2-SD\end{tabular} \\ \hline
3 & \begin{tabular}[c]{@{}c@{}}Grid representation (including inflow/outflow)\\ Graph (considering three dependencies)\end{tabular}           & \begin{tabular}[c]{@{}c@{}}3-GIO\\ 3-Gr\end{tabular}       & 3 & Sequentiality and periodicity (weekly and daily)                                                                      & 3-SWD                                               \\ \hline
\end{tabular}
\label{table:depabbr}
}
\end{table*}

\vspace{-0.2cm}
\begin{figure*}[!t]
\centering
\subfloat[Example of traffic network.]{
    \includegraphics[width=0.2\textwidth]{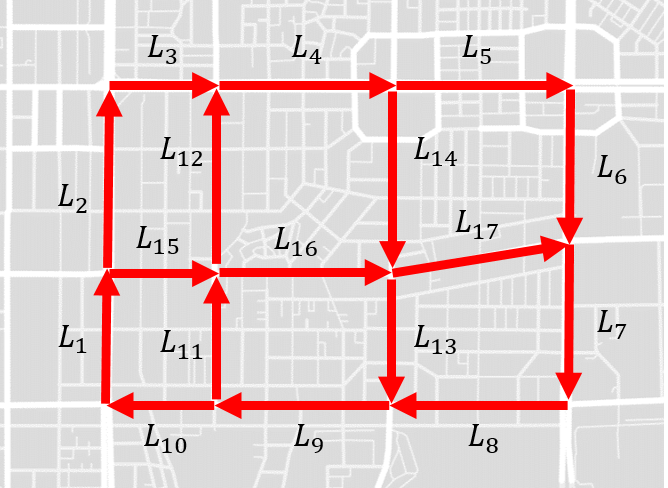}
    \label{fig:grid-a}
} \hspace{0.1cm} 
\subfloat[Grid representation applied to the input data of (a).]{
    \includegraphics[width=0.2\textwidth]{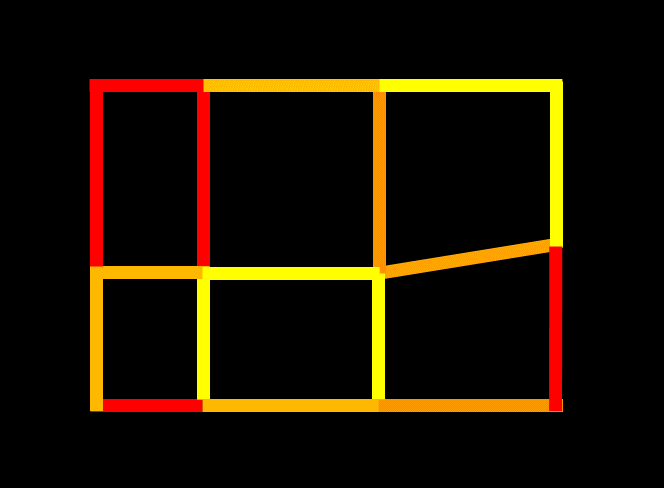}
    \label{fig:grid-b}
} \hspace{0.1cm} 
\subfloat[Example of traffic network not suitable for grid representation.]{
    \includegraphics[width=0.2\textwidth]{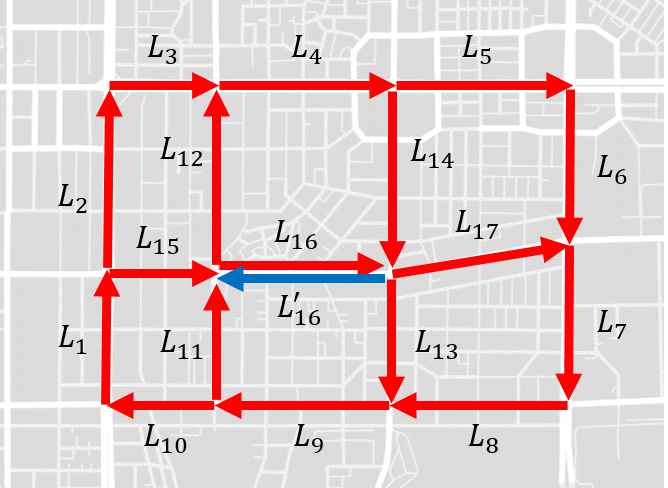}
    \label{fig:grid-c}
} \hspace{0.1cm} 
\subfloat[Example of traffic network, including surrounding area information, suitable for the graph representation method.]{
    \includegraphics[width=0.182\textwidth]{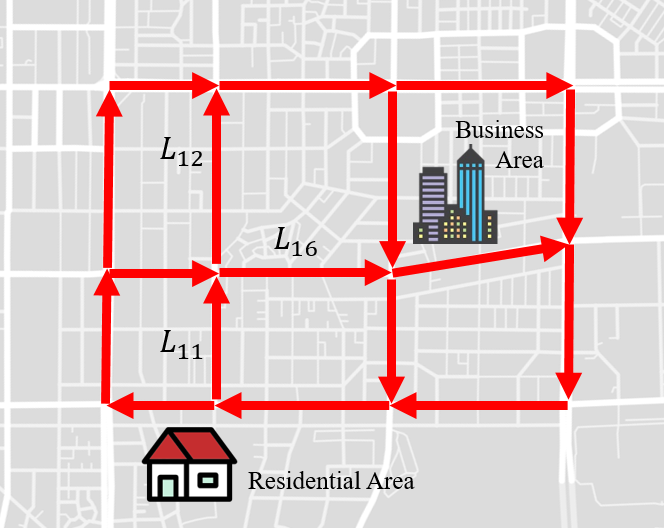}
    \label{fig:graph}
}
\caption{(a)--(c) Examples of the grid representation method, and (d) is an example of needing to apply the graph representation method. Best viewed in color.}
\label{fig:grid}
\end{figure*}

\subsubsection{Grid representation}
It is natural to regard spatial information as 2-dimensional Euclidean data consisting of latitude and longitude. Thus, without any modification, we can provide raw information to deep networks as an image-like representation. This method is also known as grid-based map segmentation or grid map. For instance, in the traffic network $L=\left\{L_1,…,L_{17}\right\}$ ($L_i$  indicates a road segment), as shown in Fig. \ref{fig:grid-a}, the final representation is the 2-d image-like representation shown in Fig. \ref{fig:grid-b}. Each pixel has a value (represented as a color in Fig. \ref{fig:grid-b}) depending on the traffic value at a specific time. In addition, recent studies have extended this method so that there are two individual channels by the flow direction (inflow/outflow). In this case, three spatial dependencies can be included in one sample.

This method is quite intuitive, but its disadvantage is the inefficiency of the representation. As shown in Fig.\ref{fig:grid-b}, most of the grid points do not match any road links, and zeros should be inserted to indicate that there are no road links. This inefficiency can worsen if higher resolution is required. 
The low resolution can also reduce the representation efficiency because several spatial units might correspond to the same grid point. In this case, we only can represent one representative value like the average value of the units. It makes the represented input data lose part of the original and essential information.
This low resolution problem can be a serious issue, particularly when a pair of road links in opposite directions should be represented, as shown in Fig. \ref{fig:grid-c}.

\subsubsection{Graphs}

Unlike image data, traffic network data have more complicated spatial dependencies (which cannot be explained only by Euclidean information) resulting from the connection order, non-regular traffic signals, and abrupt events or accidents. To consider this type of non-Euclidean dependencies, we can use graph representations. A simple example of a useful non-Euclidean spatial dependency is shown in Fig. \ref{fig:graph}, where some spatial unit labels are omitted for clarity. The labels are the same as in Figs. \ref{fig:stackedvector} and \ref{fig:grid-a}. Here, we intend to model the relation of the traffic value of $L_{11}$ to the other directly connected segments, that is, $L_{12}$ and $L_{16}$. Regarding commute hours, even though $L_{11}$ and $L_{16}$ share the driving route from the residential to the business area, $L_{12}$ does not. Thus, it would be natural to speculate that $L_{11}$ is more related to $L_{16}$ than $L_{12}$, even though both are equally proximate to $L_{11}$. 

A graph consists of nodes (vertices) and edges, and each node corresponds to a spatial unit. In a graph representation of complex pairwise information, an $N\times N$ matrix can be used, where $N$ is the number of nodes. This matrix is typically called an \textit{adjacency matrix}, and its ($i$, $j$) element represents the pairwise relationship between spatial units $i$ and $j$. Examples of complex (non-Euclidean) pairwise information include distance, connectivity, and other spatial relationships. Graph representation has recently gained increasing popularity \cite{li2017trafficspeed,yu2017trafficspeed,chu2018crowddemand,lin2018crowddemand,chai2018crowdflow,cui2019trafficspeed,iyer2018trafficspeed,lv2018trafficspeed,zhao2019trafficspeed,zhang2019trafficspeed,zhang2019trafficspeedflow,du2020crowddemand,lv2020trafficflow,lee2020trafficspeed} because the $N\times N$ matrix is an efficient means of representing spatial relationships, and the resulting DNN models have exhibited promising performance. A more detailed explanation of graph representation in traffic prediction can be found in \cite{ye2020graphsurvey}.

\subsection{Representing Temporal Dependencies}
Even though there are various methods for representing spatial dependencies, there are only \textit{two} input data representation types for temporal dependencies: \textit{sequentiality} and \textit{periodicity}. In the former, as in the case of the stacked-vector method for spatial dependencies, consecutive temporal units are stacked into a single vector to indicate that temporally closer data are related more. This is the most common method used in almost all the studies reviewed in this work. For instance, in the temporal sequence $T=\left\{t_1,…,t_m\right\}$  ($t_i$  indicates a temporal unit), the final representation is $[ V(t_1 ),V(t_2 ),…,V(t_m ) ]$, where $V(t_i)$ is the traffic value of  $t_i$ at a specific spatial unit. The size (length) of a temporal sequence can vary depending on the task.

In the latter method, temporal periodicity (e.g., hourly, daily, weekly, or monthly patterns) is represented. This method was first introduced in \cite{zhang2017crowdflow}, where two types of temporal periodicity are considered, namely, daily and weekly patterns, in addition to recent temporal sequences, that is sequentiality. For each periodicity dimension, temporal units are stacked consecutively. The type of periodicity can vary depending on the task.

\section{Deep Neural Network Techniques}
\label{sec:dnn}

The current wave of DNN evolution was initiated in 2006 by the seminal work of \cite{earlysuccess}. For the first time, it was possible to reliably train a DNN in three steps: learning a stack of Restricted Boltzmann Machines (RBMs) in a layer-by-layer manner, generating a deep autoencoder with the learned layers, and finally fine-tuning through backpropagation of error derivatives. Closely following \cite{earlysuccess}, RBMs and Deep Belief Networks (DBNs) based on stacked RBMs dominated the early applications of modern deep learning.

Once it became clear that training DNNs is possible, simpler training methods were investigated. Thorough studies were conducted on weight value initialization, optimization techniques, cost function design, and activation function design. Eventually, training DNNs for well-known tasks became straightforward even for very deep networks \cite{earlysuccess}. Multi-Layer Perceptron (MLP) \cite{firstmlp} became the default DNN architecture, whereas Convolutional Neural Networks (CNNs) \cite{firstcnnjapan,firstcnnlecun} and Recurrent Neural Networks (RNNs) \cite{firstrnn} became the method of choice for efficiently handling image (or image-like)  and sequential datasets, respectively.

With the success of MLP, CNN, and RNN, sophisticated DNN architectures and techniques were developed for various types of learning and prediction problems. Even though there are numerous such examples, we focus on the most important methods for handling STTP tasks. In the following, we categorize the DNN techniques into five different generations roughly in the order of when they first became popular. Even though such a categorization and the term \textit{generation} are not commonly used, we include them for the purpose of clearly describing the research trends in the past decade.

\subsection{Restricted Boltzmann Machines and Deep Belief Networks (First Generation)}
As the first successful DNN in \cite{earlysuccess} was based on stacked RBMs trained with an unsupervised autoencoder technique, several early STTP methods were also based on the same approach. 

\subsubsection{Restricted Boltzmann Machines}
\begin{figure}
    \centering
    \includegraphics[width=\textwidth/2]{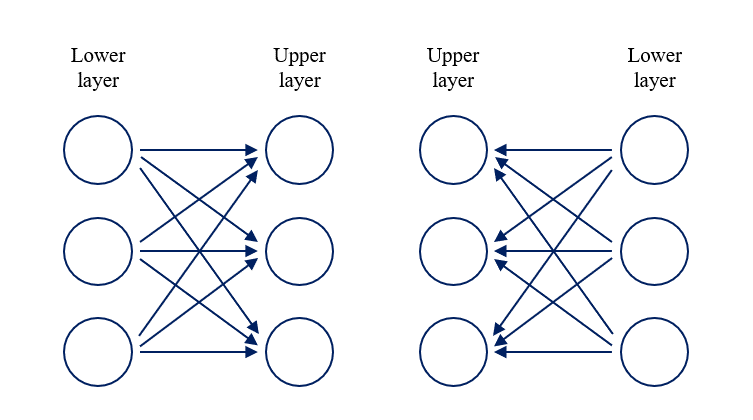}
    \caption{Schematic of Restricted Boltzmann Machine (RBM).}
    \label{fig:RBM}
\end{figure}
Boltzmann Machines (BMs) originate from statistical physics. A Boltzmann machine is a graphical model with nodes and links, where each node represents a random variable, and each link determines the interaction strength between the connected nodes. A Restricted Boltzmann Machine (RBM) is a special type of BM, where the nodes in a layer can have their links connected to the nodes in the other layer only. Thereby, update algorithms can iterate over the input and the hidden layer. An example of an RBM is shown in Fig. \ref{fig:RBM}; it can be viewed as a simple neural network as well. 

\subsubsection{Deep Belief Networks}
\begin{figure}
    \centering
    \includegraphics[width=\textwidth/2]{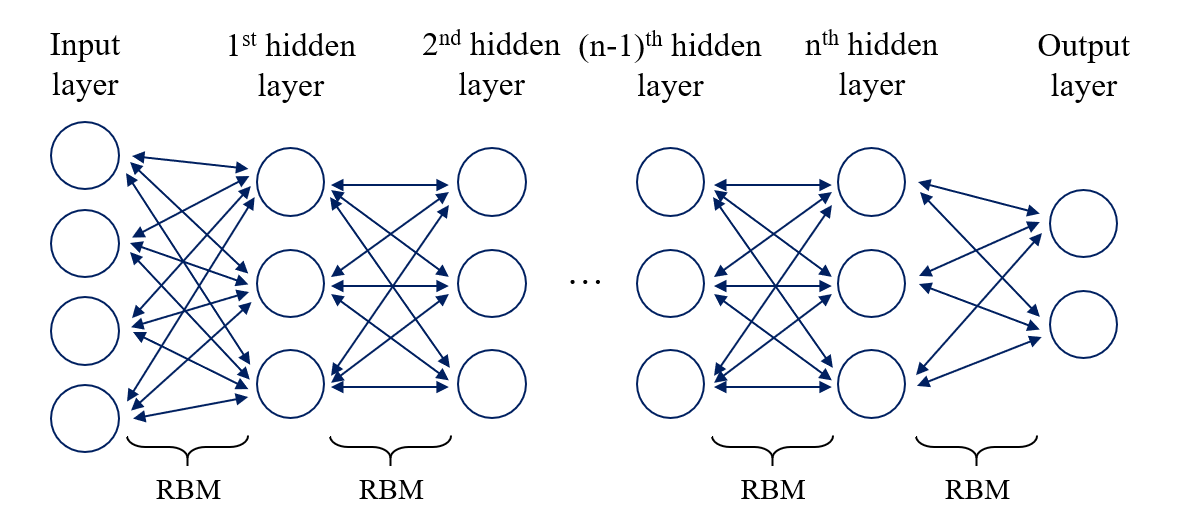}
    \caption{Schematic of Deep Belief Network (DBN).}
    \label{fig:DBN}
\end{figure}
A Deep Belief Network (DBN) is a computational model that is formed by individually learning and stacking RBMs. These networks have primarily been used as a pre-training method for supervised learning tasks; moreover, they have been used as an unsupervised feature extraction technique. Historically, DBNs are important because they were used for training early successful deep learning models. However, they gradually lost their popularity, as deep learning training was more easily performed using cross-entropy as the cost function, and Rectified Linear Unit (ReLU) as the activation function. An example of a DBN is shown in Fig. \ref{fig:DBN}.

\subsubsection{Autoencoder}
\begin{figure}
    \centering
    \includegraphics[width=\textwidth/2]{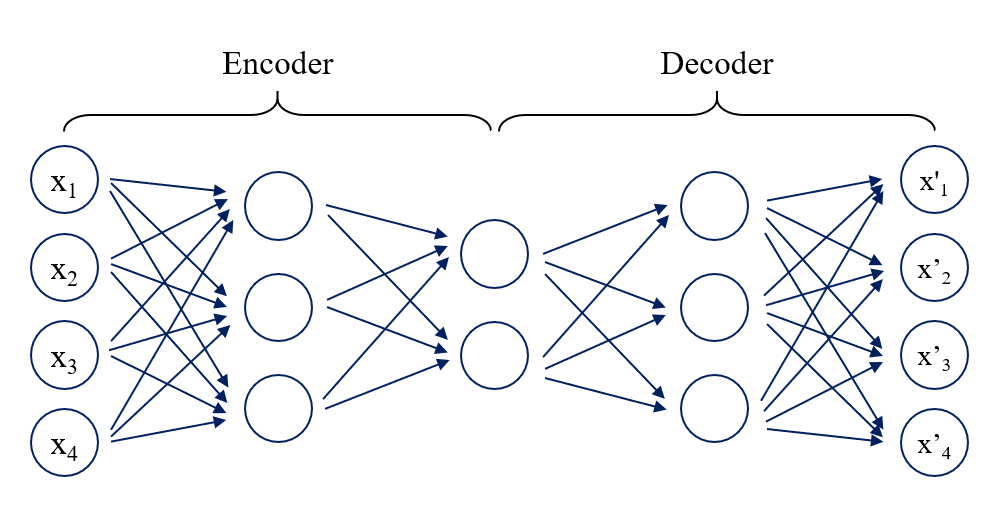}
    \caption{Schematic of autoencoder.}
    \label{fig:Autoencoder}
\end{figure}
An autoencoder is an unsupervised technique whereby an input is encoded so that it may have a small representation, and then it is decoded back to the original size. The decoded result should be as similar as possible to the original input. As shown in Fig. \ref{fig:Autoencoder}, the autoencoder architecture can be considered a feed-forward neural network with a smaller mid-layer representation and a cost function that compares the input and output. Owing to the constraints, the network is forced to learn a compact representation using a nonlinear encoder and decoder. When compared with linear methods such as principle component analysis, autoencoders can learn considerably more complicated and useful representations owing to the nonlinear nature of learning. In \cite{pcahinton}, an RBM was used for initializing each layer, and a Stacked Auto-Encoder (SAE) was used for fine-tuning the layers.   

\subsection{Multi-Layer Perceptron, Convolutional Neural Networks, and Recurrent Neural Networks (Second Generation)}
The most fundamental and yet often sufficiently powerful DNN architectures are MLP, CNNs, and RNNs. Typical traffic data form a time-series, and sequential methods, such as the classic ARIMA and Hidden Markov Model (HMM), are popular non-DNN techniques. Among the DNN techniques, an RNN is the default neural network architecture for handling time-series data with state transition in time; therefore, RNNs have been used for numerous STTP tasks. However, it has been demonstrated that CNNs can be effective even for time-series data because capturing repeated patterns in time can be more important than modeling hidden states to improve prediction performance \cite{bai2018empirical,tang2018self}.

\subsubsection{Multi-Layer Perceptron}
\begin{figure}
    \centering
    \includegraphics[width=\textwidth/2]{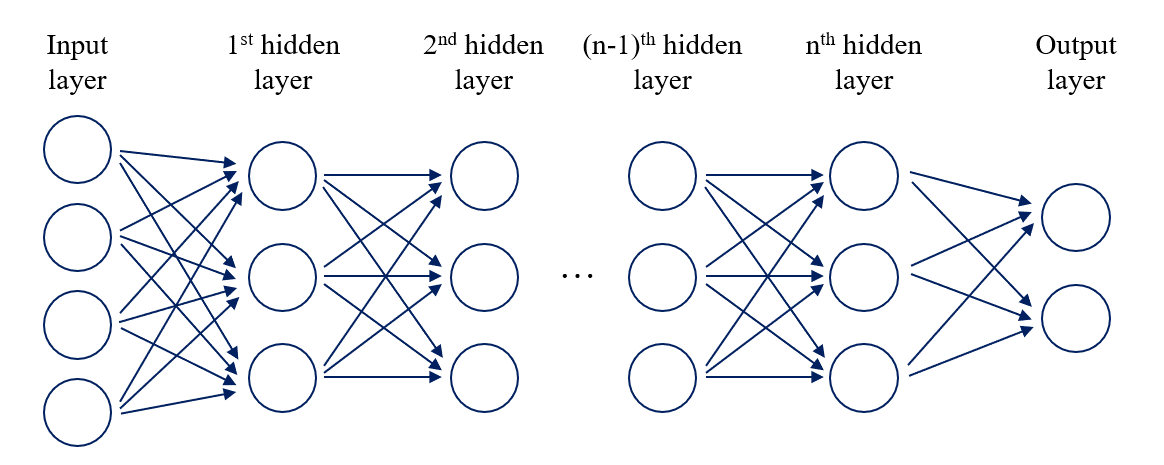}
    \caption{Schematic of Multi-Layer Perceptron (MLP).}
    \label{fig:MLP}
\end{figure}
The MLP is the most basic and fundamental architecture of modern DNNs. It was originally termed the multi-layer version of the classic perceptron \cite{firstperceptron}; it is now referred to as general feedforward network with multiple hidden layers. A schematic of the MLP is shown in Fig. \ref{fig:MLP}. As it is the most basic architecture, it usually serves as the baseline architecture of DNN algorithms.

\subsubsection{Convolutional Neural Networks}
\begin{figure}
    \centering
    \includegraphics[width=\textwidth/2]{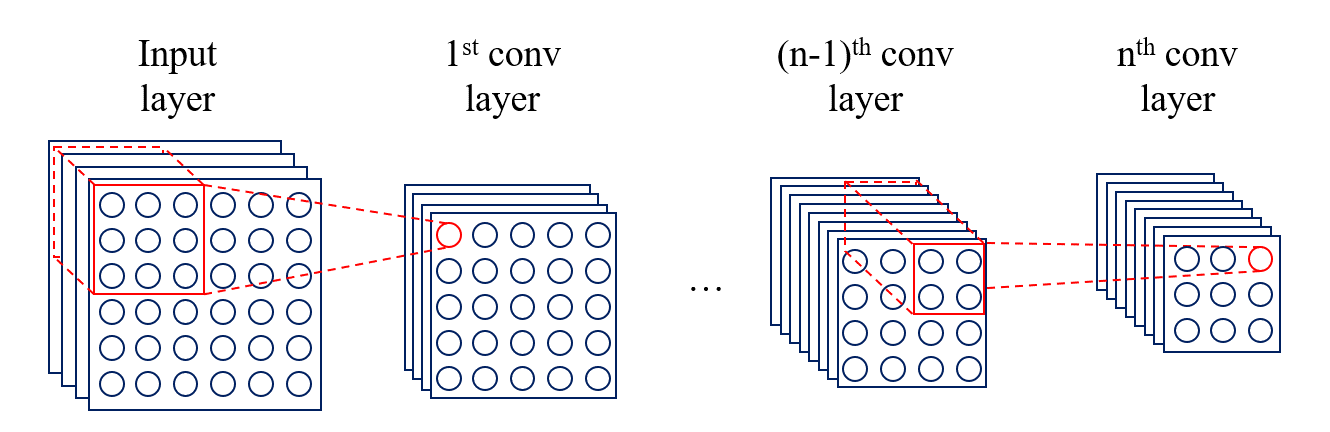}
    \caption{Schematic of Convolutional Neural Network (CNN).}
    \label{fig:CNN}
\end{figure}
In an object detection task on an image input, a shift in the object location should not affect the output of the network. Such behavior is known as shift invariance or equivariance, and CNNs are a specialized type of neural networks designed to ensure shift invariance. In addition to this property, CNNs typically require less computation power than the MLP for a given task because they share parameters and have thus sparse interactions \cite[Chapter 9]{goodfellow2016deep}. Owing to their shift invariance, CNNs have been widely used to handle image, video, and grid-like datasets. Currently, a CNN can also be used as a general building block because it is computationally efficient and provides a satisfactory performance in a variety of tasks. In some studies on STTP, CNNs have been adopted to specifically handle images or grid parts of the data \cite{wang2016trafficspeed,ma2017trafficspeed,shen2018trafficspeed}.  In several other studies, however, CNNs have been used as general building blocks. An example of a CNN is shown in Fig. \ref{fig:CNN}.

\subsubsection{Recurrent Neural Networks}
\begin{figure}
    \centering
    \includegraphics[width=\textwidth/2]{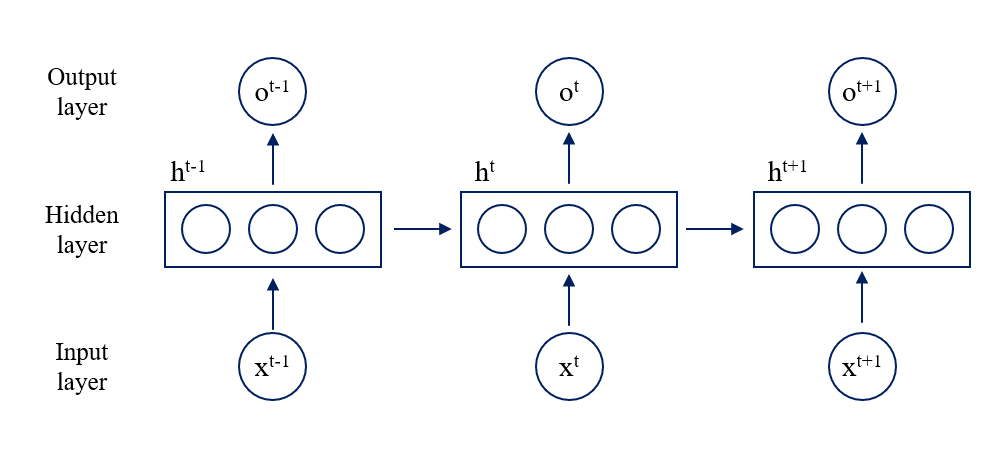}
    \caption{Schematic of Recurrent Neural Network (RNN).}
    \label{fig:RNN}
\end{figure}
Classic machine learning algorithms for handling time-sequential data include ARIMA and HMM. The latter is more advanced in that it can model latent states \cite{rabiner1989tutorial}. A considerably upgraded version of HMM is an RNN, which is far more sophisticated in modeling time dependencies than a Markov process, and can handle long-term dependencies for variable input and output size. For some applications such as machine translation, RNNs accumulate information over a long time interval and discard this information after using it. This mechanism was implemented using the concept of gating, and Long Short-Term Memory (LSTM) emerged as the most popular RNN model \cite{lstm}. An example of an RNN is shown in Fig. \ref{fig:RNN}.

Datasets for STTP usually contain spatiotemporal information with temporal information in a time-sequential form. Thus RNNs, including LSTM, have been used in numerous cases, and improvement over ARIMA, historical average, and other classic machine learning models has been reported. 

\subsection{Hybrids of CNN and RNN (Third Generation)}
\begin{figure}
    \centering
    \includegraphics[width=\textwidth/2]{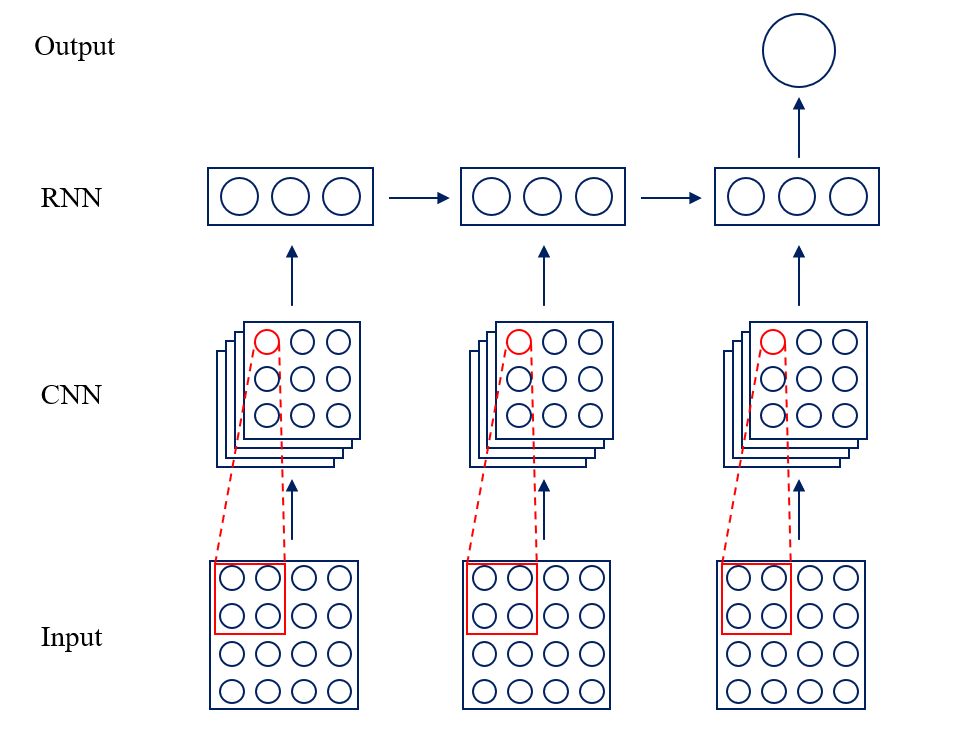}
    \caption{Example of hybrid model.}
    \label{fig:hybrid}
\end{figure}
In mainstream deep learning research, the default choice of architecture has been CNN for image data and RNN for sequential data. For instance, in \cite{vinyals2015show}, an input image is processed by a vision processing CNN and subsequently passed through a text-generating RNN. Overall, the network generates a text description of the input image. Short-term traffic predictions usually involve spatiotemporal data, where spatial information can be mapped into image or image-like format, and temporal information can be mapped into sequential format. Therefore, it is natural to use a hybrid network, where a CNN is applied to the spatial part, and an RNN to the temporal part. In practical implementations, however, a strong mapping between data and architecture types is not necessary because both the CNN and RNN can be used for general modeling of any data type. Mapping is recommended but is not necessarily the best-performing model. Fig. \ref{fig:hybrid} shows an example of hybrid model.

\subsection{Graph Convolutional Networks (Fourth Generation)}
\vspace{-0.2cm}
\begin{figure}[!h]
\centering
\subfloat[2D convolution.]{
    \includegraphics[width=0.18\textwidth]{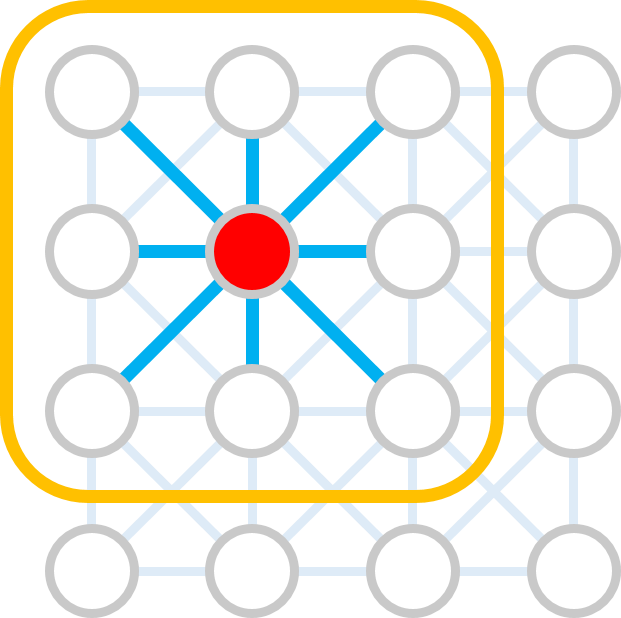}
    \label{fig:graphs-a}
} \hspace{0.2cm}
\subfloat[Graph convolution.]{
    \includegraphics[width=0.22\textwidth]{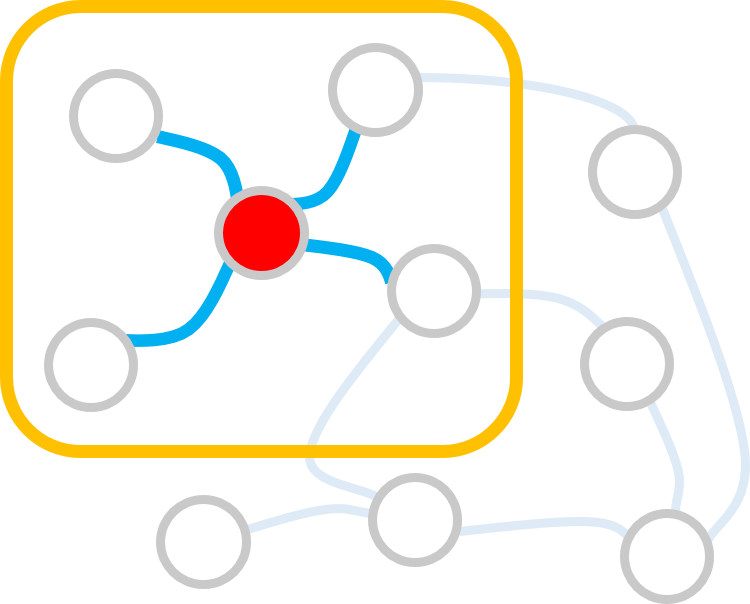}
    \label{fig:graphs-b}
}
\caption{Typical 2D convolution and a graph convolution. Best viewed in color. (a) 2D convolution. (b) Graph convolution. (Figures were adapted from Fig. 1 of \cite{wu2020comprehensive} and presented in Fig. 3 of \cite{lee2020trafficspeed}.)}
\label{fig:graphs}
\end{figure}
As explained in Section 2, a graph is a powerful means of representing spatial information, such as on-path proximity, $k$-hop connectivity, and driving directions. In a typical graph- and DNN-based modeling process, spatial relationships that can be best represented in a graph are first identified. Then, the pair-wise information among the $N$ nodes is summarized into an $N \times N$ matrix for each graph relationship. A matrix that represents spatial interdependencies is called adjacency matrix, and its $(i,j)$ element represents the pairwise relationship between node $i$ and node $j$. For instance, the most widely used adjacency matrix is the distance matrix, where its $(i,j)$ element contains the on-path distance information between the road segment $i$ and road segment $j$. A graphical explanation is shown in Fig. \ref{fig:graphs}. Once the adjacency matrices are generated, they are used as the design elements of a Graph Convolutional Network (GCN). Even though early GCNs were introduced to bridge spectral graph theory and DNNs \cite{bruna2013spectral}, they have evolved into simpler architectures with high computational efficiency. These networks as DNN models, however, have a sufficiently large capacity and the potential for capturing complex spatial relationships despite the simplification.

Graph convolutional networks need not explicitly use convolutional layers; however, the term \textit{convolutional} is used because of the manner in which information is packed into adjacency matrices. In Fig. \ref{fig:graphs}, a typical 2D convolution is shown together with a graph convolution. In both, the information of neighboring nodes, that is, the nodes inside the filter (orange box), is used for processing the information of the node of interest (shown in red). In 2D convolution, defining the filter shape and size is straightforward because the underlying data are well structured, where all the neighboring nodes connected with blue lines are located inside the filter. However, this is not case for graph convolution. Unlike grid data, traffic network data are not well structured, and the size of the neighborhood of a node can vary. Instead of defining a small and common filter for all nodes, an $N \times N$ matrix is used as a spatial filter, where neighborhood information is collectively represented in the matrix. In row $i$ of the matrix, all the neighboring nodes connected with blue lines from node $i$ are represented with non-zero weight values, whereas the others are marked with zero.

\subsection{Other Advanced Techniques (Fifth Generation)}
The most fundamental deep learning architectures are MLP, CNN, and RNN, and a variety of techniques can be combined to augment their capabilities, as for example GCNs. We now discuss widely used techniques in STTP, as well as a few other techniques that have been considered in deep learning studies, but have not been heavily applied in transportation research. 

\subsubsection{Transfer Learning}
Transfer learning is a machine learning technique in which knowledge is obtained by solving a problem, and the acquired knowledge is used for solving a related but different problem. For instance, we can consider learning a traffic prediction function using the data of a metropolitan city $A$, and then apply the learnt prediction function to another metropolitan city $B$.  According to \cite{pan2009survey}, transfer learning can be formally defined as follows: \\

(Transfer learning) Given a source domain $D_{S}$ and learning task $T_{S}$, and a target domain $D_{T}$ and learning task $T_{T}$, transfer learning is aimed at improving the learning of the target predictive function $f_{T}(\cdot)$ in $D_{T}$ using the knowledge in $D_{S}$ and $T_{S}$, where $D_{S} \neq D_{T}$ or $T_{S} \neq T_{T}$.\\

There are a variety of scenarios and methods for transfer learning, and a survey can be found in \cite{pan2009survey}. For deep learning, the most typical scenario is when a sufficient amount of labeled data is available for the source problem but not for the target problem. In this case, fine-tuning can be applied, in which a network is trained by solving the source problem. Subsequently, the network weights are updated using the data and task in target domain. 
Considering that several factors of traffic environments, such as traffic lights, frequent road topologies, ramps to expressways or highways, and commute time, are common, transferring knowledge from one traffic dataset to another can be an effective approach. The transfer of knowledge can be applied in the space and the time dimension (or both).

\subsubsection{Meta Learning}
\begin{figure}
    \centering
    \includegraphics[width=\textwidth/2]{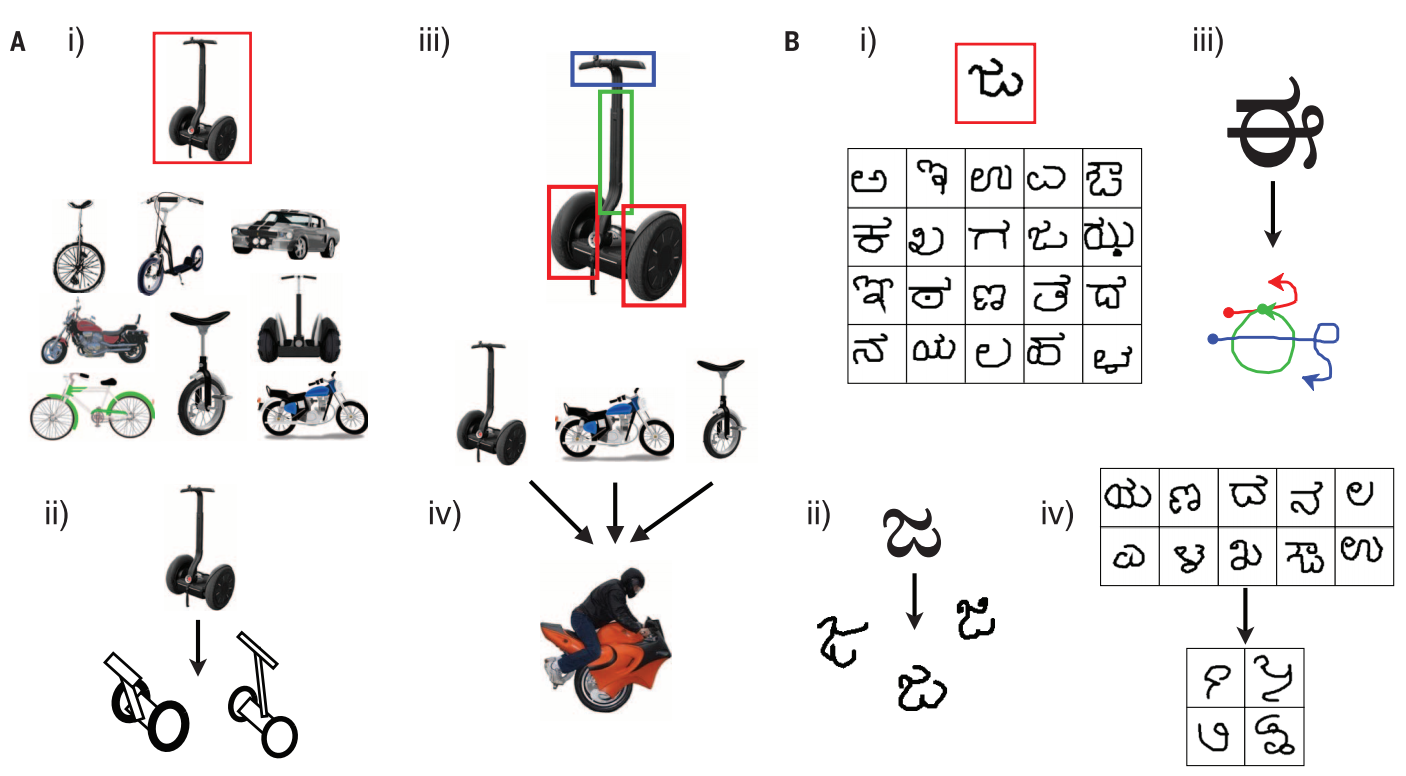}
    \caption{Rich concepts can be learnt from limited data (Figure from \cite{lake2015human}).}
    \label{fig:meta_learning}
\end{figure}
``Fast mapping'' (i.e., the meaningful generalizations that can be made from one or a few examples of a novel word) is arguably the most remarkable feat of human word learning \cite{carey1978acquiring}, and it has become popular in machine learning by a few recent studies, including \cite{lake2015human}. Fig. \ref{fig:meta_learning} is provided in \cite{lake2015human} and shows human capabilities that do not require large data. We first consider the example of a personal transporter on the left side $A$ of Fig. \ref{fig:meta_learning}. Even at first sight, humans can deduce that (i) this object resembles a bicycle, motorcycle, vehicle, etc. through its appearance, (ii) it can be expressed in many different forms without even knowing what it is, (iii) its components can be separated by handles, wheels, supports, etc., and (iv) it can be integrated with similar examples to generate the concept of a new transporter. On the right side $B$ of Fig. \ref{fig:meta_learning}, a similar example is shown for Omniglot \cite{lake2015omniglot}, a data set of 1623 handwritten characters from 50 writing systems. In deep learning, numerous methods have been developed since 2015, where previously learnt knowledge is used for successfully completing a new but related task using only a small number of examples. This is termed meta learning and can be considered a type of transfer learning.  

\subsubsection{Reinforcement Learning}
Reinforcement learning is fundamentally different from typical supervised or unsupervised learning in that learning is performed by interacting with the environment. More precisely, it focuses on goal-directed learning, where a mapping between the situation and the action should be learned through the interactions between the agent and the environment. Even though reinforcement learning is relatively old, it has become highly popular since the historical win over human experts in the game of ``GO'' \cite{silver2016mastering} \cite{silver2017mastering}. By integrating DNNs into reinforcement learning, the modeling accuracy of the policy and the value function was dramatically improved, where policy is defined as the mode of behavior of the learning agent at a given time, and value is defined as the total amount of rewards that an agent can expect to accumulate in the long run. In transportation, reinforcement learning can be effective if action and reward can be clearly defined. For instance, it can be useful for bike repositioning, drone path control, or autonomous driving policy control as well as properly defined STTP problems.

\subsubsection{Attention}
\begin{figure}
    \centering
    \includegraphics[width=\textwidth/2]{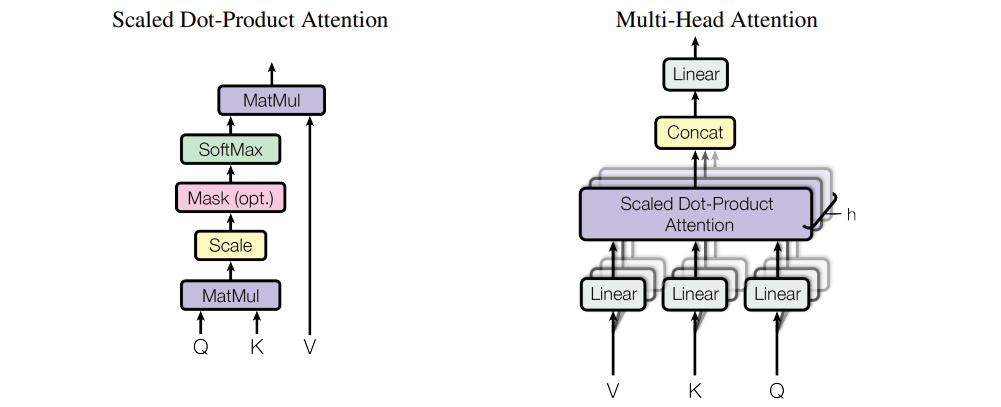}
    \caption{Scaled dot-product attention and multi-head attention (Figure from \cite{vaswani2017attention}).}
    \label{fig:attention}
\end{figure}
The attention mechanism is based on a simple heuristic: Humans attend to a specific part when processing a large amount of information, such as text of an image. Even though its first appearance in deep learning was primarily as a component for machine translation \cite{cho2014learning}, it has now become the most important standard part in state-of-the-art Natural Language Processing (NLP) models. The Transformer architecture \cite{vaswani2017attention} is a DNN model that is solely based on the attention mechanism, and the technique of bidirectional encoder representations from transformers \cite{devlin2018bert} is an improved transformer architecture with state-of-the-art performance in a wide range of NLP tasks. Inspired by this remarkable performance, researchers are attempting to use attention in other application areas as well. Fig. \ref{fig:attention}, which is from \cite{vaswani2017attention}, shows the mechanism of self-attention.

\subsubsection{Other Possibilities}
Deep neural networks are unique in that they have millions of parameters enabling complex function modeling. They use the simple but highly effective training method of stochastic gradient descent, whereby important features and patterns may be automatically discovered. Great progress has been made in the past ten years, and the intensity of deep learning research is expected to remain high in the foreseeable future. Several new techniques useful for STTP are expected to be developed. Among the currently available techniques, generative models, such as variational autoencoders \cite{kingma2013auto,pu2016variational} and generative adversarial networks \cite{goodfellow2014generative}, may be directly applicable in STTP. In addition, remarkable progress has recently been made in self-supervised learning \cite{lan2019albert,chen2020simple}, which is expected to be quite useful in STTP problems, where a large amount of unlabeled data is available. 

\section{Application Areas}
\label{sec:appl}

A summary of the surveyed papers are shown in Table \ref{table:all} where the papers are organized in chronological order considering the DNN technique generation and publication year. However, we provide a summary of the selected papers according to application area below. We classified the application areas depending on the entity type and target measure.

\subsection{Crowd Flow Prediction}
Crowd flow refers to the number of people in a pre-defined region in a given time period. As there are a variety of transportation types, such as bikes and subways, or simply walking, there are no clear spatial restrictions on the moving paths. Accordingly, in most studies\cite{zhang2017crowdflow,chen2018crowdflow,wang2018crowdflow,zhang2018crowdflowcnn,lin2019crowdflow,rong2019crowdflow,zhang2019crowdflow,ma2018crowdflow,zonoozi2018crowdflow,ren2019crowdflow,li2019crowdflow}, the regions are defined first, and then grid representation is used for spatial dependencies. In some cases, crowd flow can be divided into inflow and outflow. 

In most surveyed crowd flow prediction studies, the spatial dependency is considered through grid representation based on coordinates and flow direction (i.e., inflow/outflow), and the temporal dependency is considered through sequentiality and periodicity (daily, weekly). Reference \cite{zhang2017crowdflow} was the first study to use this type of representation for crowd flow prediction. Specifically, input data are represented as 2-channel and 2-dimensional images including three spatial dependencies: Each channel corresponds to inflow and outflow and each dimension corresponds to latitude and longitude. The divided region refers to a pixel in the image, and each pixel has a crowd flow value per timestamp. In addition, to consider the sequentiality and two periodicities, three individual networks are constructed for each time period, and a fusion network is used to combine them. Later, several studies improved \cite{zhang2017crowdflow} in various ways. For example, in \cite{zonoozi2018crowdflow}, the network from a CNN is transformed into a hybrid network (ConvGRU), and in \cite{wang2018crowdflow}, Gaussian noise is introduced into the hidden units.

In \cite{chai2018crowdflow}, the spatial dependency is represented as a multi-graph based on three types of non-Euclidean spatial dependencies of the bike dataset. For all pairs of bike stations, the distance graph is defined as the symmetrical distance, the interaction graph is defined as the number of ride records, and the correlation graph is defined as the Pearson correlation coefficient of past records. Each graph is separately learnt by an individual CNN. Then, an encoder–decoder LSTM is constructed to consider temporal correlation. 

Reference \cite{wang2018crowdflow} is aimed at effectively transferring knowledge from a data-rich source city to a data-scarce target city. The matching function is defined based on history and auxiliary data, and then the prediction network referenced from previous works, such as ST-ResNet \cite{zhang2017crowdflow}, is fine-tuned.

\subsection{Crowd Demand Prediction (Taxi, Bike, Ridesharing)}
Crowd demand refers to the number of people with pickup or dropoff demands, such as taxi, bike-sharing, or ridesharing, in a pre-defined region in a given time period. Accurate demand prediction can lead to an efficient disposition of supplies. As in the case of crowd flow, the entity type of which is humans, there is no spatial restriction; thus, in most studies\cite{huang2017crowddemand,ke2017crowddemand,liao2018crowddemand,wang2018crowddemand,yao2018crowddemand,zhou2018crowdflow,yao2018crowddemand,yao2019crowddemand,he2019crowddemand,zhou2020crowddemand,daganzo2019general}, the regions are defined first, and then grid representation is used for the spatial dependency. 

In most surveyed crowd demand prediction studies, the spatial dependency is considered through grid representation based on coordinates or graphs with non-Euclidean dependencies, and the temporal dependency is considered through sequentiality only. In \cite{yao2018crowddemand} a hybrid network was constructed to simultaneously consider three different perspectives: a CNN for the spatial perspective, LSTM for the temporal perspective, and dynamic time warping with the embedding vectors resulting from the large-scale information  network embedding algorithm for the semantic perspective.

In \cite{lin2018crowddemand}, five types of non-Euclidean spatial dependencies of the bike demand dataset were compared: spatial distance, demand matrix, average trip duration as binary, demand correlation, and fully trainable matrix. The fully trainable matrix exhibited the best performance. In \cite{geng2019crowddemand}, a multi-graph including three types of non-Euclidean spatial dependencies is defined. By representing regions as grids, the neighborhood of a region is defined based on the spatial proximity in binary, functional similarity is defined as the similarity between point-of-interest vectors, and transportation connectivity is defined based on the connectedness through motorways, highways, or public transportation, such as subway systems. For each graph, graph convolution is used individually, and then the outputs are fused. To model temporal correlation, a contextual gated recurrent neural network was further proposed. This network augments an RNN with a contextual-aware gating mechanism to re-weight different historical observations. 

\clearpage
\midsepremove
{
\scriptsize
\begin{center}
\captionsetup{width=\textwidth}

\setlength{\LTleft}{-20cm plus -1fill}
\setlength{\LTright}{\LTleft}


\end{center}
}
\midsepdefault
\normalsize

\subsection{Traffic Flow Prediction}

Traffic flow refers to the number of vehicles passing through a spatial unit, such as a road segment or traffic sensor point, in a given time period. Accurate prediction assists in preventing traffic congestion and managing traffic conditions in advance. Unlike humans, vehicles can move only through pre-defined roads; accordingly, stacked vector and graphs are often used for the input data representation. 

In one of the first STTP studies, SAE networks were trained \cite{lv2014trafficflow}. Stacked vector and sequentiality were used for the input data representation. In \cite{zhang2019trafficspeedflow}, the spatial dependency is represented as a graph based on the temporal correlation coefficient using historical traffic observations to capture heterogeneous spatial correlations. To calculate the correlation coefficient, min-max normalization is first used to calculate the coefficient so that  spatial heterogeneity may be eliminated from capacity or speed limits. Then, the daily periodicity is also removed using the $z$-score transformation because it may result in strong temporal auto-correlation. Finally, the network is trained including graph convolution and residual LSTM.

In \cite{pan2019trafficflow}, ST-MetaNet was proposed. Its input data are represented as graphs based on connecting patterns or distance. A sequence-to-sequence architecture consisting of an encoder for learning historical information, and a decoder for step-by-step prediction are employed. Specifically, the encoder and decoder have the same network structure: a meta graph attention network to capture spatial correlations, and a meta recurrent neural network to consider temporal correlations.

\subsection{Traffic Speed Prediction}

Traffic speed refers to the average speed of vehicles passing through a spatial unit, such as a road segment or traffic sensor point, in a given time period. As in the case of traffic flow, the entity type of which is vehicles, stacked vectors and graphs are often used for the input data representation. 

Traffic speed prediction can be divided into two types according to the traffic network considered. The first type considers only freeways. As freeways have a relatively simple structure with a few traffic signals or on/off-ramps, prediction accuracy can be improved by simply increasing the temporal resolution of the dataset. Hence, in the case of traffic speed prediction for freeways, temporal dependency has been emphasized more than spatial dependency. Therefore, input data are often represented by simple stacked-vector methods. In \cite{ma2017trafficspeed}, the input data are represented as a 2-d matrix combining the spatial and temporal dependencies. Each column of the matrix corresponds to the stacked vector in the connection order, and each row represents the temporal sequentiality. 

In recent studies, more complex traffic networks, such as urban networks, have been considered. These networks may have considerably more complicated connection patterns and abrupt changes. Thus, more sophisticated spatial dependency representation methods, such as graphs considering proximity or pairwise distance, have been introduced. In \cite{li2017trafficspeed}, bidirectional diffusion convolution is performed on the graph to capture spatial dependency, and a sequence-to-sequence architecture with gated recurrent units is used to capture temporal dependency. Input graphs are defined based on the shortest path distance between traffic sensors. Later, in \cite{yu2017trafficspeed}, the training time was shortened by replacing recurrent layers by convolutional layers. Recently, in \cite{lee2020trafficspeed}, the input data are represented as multi-graphs based on three types of spatial dependencies: distance, direction, and positional relationships. The graph weights are also modified using simple partition filters.

Reference \cite{ma2018trafficspeed} proposed a network consisting of a capsule network with grid representation for the input data and nested LSTM to consider temporal sequentiality. In \cite{zhang2019trafficspeed}, a distance-based graph is first transformed into vectors. Then, a spatial embedding is trained to encode vertices into vectors that preserve the graph structure information. The attention network is manipulated for the final prediction.

\subsection{Traffic accidents/congestion prediction}

Unlike the aforementioned applications corresponding to the regression problem, traffic accident/congestion prediction is related to the classification problem. For accident prediction, some studies predict whether an accident occurs through a binary classification, and others predict the injury levels resulting from accidents through a multi-class classification. For congestion prediction, the congestion level is determined by the specific traffic flow or traffic speed value. Typical studies predict congestion through a binary classification; congestion is defined based on an average speed less than 20 km/h.

Reference \cite{zhang2019trafficaccident} proposed transferring the prediction network from the source area, which has the ground-truth traffic accident reports, to the target area, which does not. Traffic-accident-related features, such as accident location and time, are first extracted from unstructured social sensing data (e.g., tweets). Then, a transformation network and a loss function are defined to consider the discrepancy between the source and target distributions. Finally, all networks are trained using the adversarial method. 

In \cite{sun2019trafficcongestion}, the congestion level is predicted from a CNN and an RNN, which are individually trained. The input data for the CNN are represented by stacked vectors and temporal sequentiality, whereas the input data for the RNN are represented by the temporal sequentiality of the average speed over multiple spatial units. Finally, the traffic congestion levels are classified according to the output of the CNN and RNN.

\section{Public Datasets}
\label{sec:dataset}

A major problem in STTP studies is the lack of standard benchmark datasets \cite{veres2019survey}. Benchmark datasets can greatly facilitate research because findings can be easily compared and integrated, whereas the lack of benchmark datasets may result in data-specific rather than generalizable studies. Accordingly, we summarize publicly available often-used datasets in Table \ref{table:dataset}. Dataset availability was determined at the time of writing.

\begin{table*}[ht!]
\centering
\caption{Summary of Public Datasets. It should be noted that dataset names are not official.}
\resizebox{\textwidth}{!}{%
\begin{tabular}{ccccll}
\hline
\multirow{2}{*}{\textbf{Dataset}} & \multirow{2}{*}{\textbf{\begin{tabular}[c]{@{}c@{}}City,\\ Country\end{tabular}}} & \multirow{2}{*}{\textbf{Duration}}                                     & \multirow{2}{*}{\textbf{\begin{tabular}[c]{@{}c@{}}Time\\ Resolution\end{tabular}}} & \multicolumn{1}{c}{\multirow{2}{*}{\textbf{Spatial Coverage}}}                                                                                 & \multicolumn{1}{c}{\multirow{2}{*}{\textbf{Data Type (Examples)}}}
\\
                                  &                                                                                   &                                                                        &                                                                                     & \multicolumn{1}{c}{}                                                                                                                           & \multicolumn{1}{c}{}                                                                                                                                                                        \\ \hline
\textbf{PeMS} \cite{dataset:pems}                     & \begin{tabular}[c]{@{}c@{}}California,\\ USA\end{tabular}                         & 2001$\sim$                                                                  & 5min                                                                                & \begin{tabular}[c]{@{}l@{}}Nearly 40,000 individual detectors, spanning the freeway \\ system across all major metropolitan areas\end{tabular} & \begin{tabular}[c]{@{}l@{}}Traffic detectors, incidents, lane closures, toll tags, census traffic counts, \\ vehicle classification, weight-in-motion, roadway inventory, etc.\end{tabular} \\ \hline
\textbf{QTraffic} \cite{dataset:qtraffic}                 & \begin{tabular}[c]{@{}c@{}}Beijing,\\ China\end{tabular}                          & \begin{tabular}[c]{@{}c@{}}4/1 2017\\ $\sim$5/31 2017\end{tabular}          & 15min                                                                               & \begin{tabular}[c]{@{}l@{}}15,073 road segments covering \\ approximately 738.91 km (6th ring road)\end{tabular}                                & \begin{tabular}[c]{@{}l@{}}Traffic speed, user queries (timestamp, starting/destination location), \\ Road network information (width, direction, length, etc.)\end{tabular}          \\ \hline
\textbf{TaxiNYC} \cite{dataset:taxinyc}                  & \begin{tabular}[c]{@{}c@{}}New York,\\ USA\end{tabular}                           & 2009$\sim$          & \begin{tabular}[c]{@{}c@{}}-\\ (Per trip)\end{tabular}                              & \begin{tabular}[c]{@{}l@{}}Including Bronx, Brooklyn, Manhattan, \\ Queens, Staten Island\end{tabular}                                         & \begin{tabular}[c]{@{}l@{}}Pick-up and drop-off date/time/location, trip distance, \\ rate/payment type, passenger count\end{tabular}                                         \\ \hline
\textbf{TaxiBJ} \cite{dataset:taxibj}                  & \begin{tabular}[c]{@{}c@{}}Beijing,\\ China\end{tabular}                          & \begin{tabular}[c]{@{}c@{}}7/1 2013\\ $\sim$4/30 2016\end{tabular}          & 30 min                                                                               & Over 34,000 taxis                                                                                                                              & Trajectory, meteorological data                                                                                                                                                              \\ \hline
\textbf{BikeNYC} \cite{dataset:bikenyc}                 & \begin{tabular}[c]{@{}c@{}}New York,\\ USA\end{tabular}                           & 5/27 2013$\sim$                                                                  & \begin{tabular}[c]{@{}c@{}}-\\ (Per trip)\end{tabular}                                                                               & Over 6,800 bikes                                                                                                                               & \begin{tabular}[c]{@{}l@{}}Trip duration, start/end station ID/time,\\ bike ID, user type, gender, year of birth\end{tabular} \\ \hline
\textbf{BusNYC} \cite{dataset:busnyc}                  & \begin{tabular}[c]{@{}c@{}}New York,\\ USA\end{tabular}                           & 2014                                                                   & 10min                                                                               & All public buses                                                                                                                               & \begin{tabular}[c]{@{}l@{}}Coordinates (latitude, longitude), timestamp of transmission, \\ distance to the next bus stop\end{tabular} \\ \hline
\textbf{WebTRIS} \cite{dataset:webtris}                 & England                                                                           & 4.2015$\sim$                                                                & 15min                                                                               & \begin{tabular}[c]{@{}l@{}}All motorways and 'A' roads managed \\ by the Highways England (Strategic Road Network)\end{tabular}                & \begin{tabular}[c]{@{}l@{}}Journey time, traffic flow, longitude/latitude/length of link, \\ flow quality, quality index\end{tabular}  \\ \hline
\textbf{DRIVENet} \cite{dataset:drivenet}                    & \begin{tabular}[c]{@{}c@{}}Seattle,\\ USA\end{tabular}                            & 2011                                                                   & 1min                                                                                & 323 stations, 85 miles                                                                                                                         & \begin{tabular}[c]{@{}l@{}}Pedestrian travels, public transit data, traffic flow, \\travel time, safety performance, incident induced delay \end{tabular} \\ \hline
\textbf{Porto} \cite{dataset:porto}                   & \begin{tabular}[c]{@{}c@{}}Porto,\\ Portugal\end{tabular}                         & \begin{tabular}[c]{@{}c@{}}7/1 2013\\ $\sim$6/30 2014\end{tabular}          & \begin{tabular}[c]{@{}c@{}}-\\ (Per trip)\end{tabular}                              & \begin{tabular}[c]{@{}l@{}}442 taxis, 420,000 trajectories,\\ 16735m*14389m\end{tabular}                                                       & call type, timestamp, day type, GPS coordinates \\ \hline
\textbf{Didi} \cite{dataset:didi}            & \begin{tabular}[c]{@{}c@{}}Chengdu,\\ China\end{tabular}                          & \begin{tabular}[c]{@{}c@{}}11/1 2016\\ $\sim$11/30 2016\end{tabular}                                                               & 2$\sim$4secs                                                                                 & \begin{tabular}[c]{@{}l@{}}5476 geographical blocks \\ (each block size = 1 $km^2$)\end{tabular}                                                    & Order ID, start/end billing time, pick-up/drop-off longitude/latitude  \\ \hline
\textbf{BikeDC} \cite{dataset:bikedc}                  & \begin{tabular}[c]{@{}c@{}}Washington,\\ USA\end{tabular}                         & 9/20 2010 $\sim$                                                                  & \begin{tabular}[c]{@{}c@{}}-\\ (Per trip)\end{tabular}                              & \begin{tabular}[c]{@{}l@{}}Over 500 stations across 6 jurisdictions,\\ 4,300 bikes\end{tabular}                                                & Duration, start/end date, start/end station, bike number, member type                \\ \hline
\textbf{BikeCHI} \cite{dataset:bikechi}                 & \begin{tabular}[c]{@{}c@{}}Chicago,\\ USA\end{tabular}                            & 6/27 2013$\sim$                                                                  & \begin{tabular}[c]{@{}c@{}}-\\ (Per trip)\end{tabular}                              & \begin{tabular}[c]{@{}l@{}}580 stations across Chicagoland,\\ 5,800 bikes\end{tabular}                                                         & Trip start/end time/station, rider type (member, single ride, explore pass)                    \\ \hline
\textbf{HK} \cite{dataset:hk}                      & \begin{tabular}[c]{@{}c@{}}Hong Kong,\\ Hong Kong\end{tabular}                    & 12/28 2015$\sim$                                                            & 2 min                                                                                & four regions in Hong Kong                                                                                                                         & \begin{tabular}[c]{@{}l@{}}Link ID, region, road type, road saturation level (good/bad/average), \\ speed, date\end{tabular}    \\ \hline
\textbf{Uber} \cite{dataset:uber}                    & \begin{tabular}[c]{@{}c@{}}New York,\\ USA\end{tabular}                           & \begin{tabular}[c]{@{}c@{}}4/1 2014$\sim$9/30 2014,\\ 1/1 2015$\sim$6/30 2015\end{tabular} & \begin{tabular}[c]{@{}c@{}}-\\ (Per trip)\end{tabular}                              & Over 18.8 million pickups                                                     & Date/time, latitude, longitude \\ \hline
\textbf{TOPIS} \cite{dataset:topis}                   & \begin{tabular}[c]{@{}c@{}}Seoul,\\ South Korea\end{tabular}                      & 2014.1$\sim$                                                                & 5 min, 1 h                                                                         & 1,153 sensors and over 70,000 taxis(GPS)                                                                                                       & \begin{tabular}[c]{@{}l@{}}Date/time, coordinates, road type, region type, \\ average speed/flow, number of lanes\end{tabular} \\ \hline
\textbf{BusCHI} \cite{dataset:buschi}              & \begin{tabular}[c]{@{}c@{}}Chicago,\\ USA\end{tabular}                            & \begin{tabular}[c]{@{}c@{}}8/2 2011\\ $\sim$ 5/3 2018\end{tabular}                   & 10 min                                                                               & \begin{tabular}[c]{@{}l@{}}Buses on arterial streets in real-time, \\ about 1,250 road segments covering 300 miles\end{tabular}                & Time, segment ID, bus count, reading count, speed \\ \hline
\end{tabular}
}
\label{table:dataset}
\end{table*}

The most widely used dataset for traffic flow and speed prediction area is the Caltrans performance measurement system, called the \textit{PeMS} dataset (used in 27 papers in this survey).  This dataset is managed by the California Department of Transportation. All data are available from 2001 to the present. The \textit{PeMS} dataset is based on nearly 40,000 individual detectors, spanning the freeway system across all major metropolitan areas of California. The final data resolution is 5 min, aggregated from 30 s raw measurements. The \textit{PeMS} dataset is provided as a \textit{csv} file, including information on traffic detectors, incidents, lane closures, toll tags, census traffic counts, vehicle classification, weight-in-motion, roadway inventory, etc. This dataset includes abundant spatiotemporal information; however, it is limited to simple freeways only. Thus, the insights from the studies based on this dataset may not be directly used for prediction in complex urban networks. 

For crowd flow and demand prediction, there are three often-used datasets: \textit{BikeNYC}, \textit{TaxiBJ}, and \textit{TaxiNYC}. The data in \textit{BikeNYC} (used in 18 papers in this survey) have been collected from CitiBike, New York, USA, since 2013. The data are from more than 6,800 membership users in New York City, with a time resolution of 1 h. The \textit{BikeNYC} dataset includes various information, such as trip duration, starting/ending station, starting/ending time, bike ID, user type, gender, and year of birth. The dataset can be downloaded in \textit{csv} format.

The data in \textit{TaxiBJ} (used in nine papers in this survey) have been collected from Beijing, China, for over three years. These data are provided in \cite{zhang2017crowdflow}. They have a time resolution of 30 min. The dataset includes trajectory and meteorology data from over 34,000 taxis in Beijing. It is primarily used for crowd flow research.

The data in \textit{TaxiNYC} (used in nine papers in this study) were collected from NYC Taxi and Limousine Commission by technology providers authorized under the Taxicab \& Livery Passenger Enhancement Programs  from January 2009 to June 2019. This dataset contains taxi trip records including pick-up/drop-off time and location, trip distance, itemized fare, rate type, payment type, and driver-reported passenger count. It is primarily used for demand prediction.

\section{Discussion and Conclusion}
\label{sec:discussion}

We reviewed recent DNN-based STTP research according to input data representation methods, DNN techniques, application areas, and dataset  availability. Herein, we suggest challenges and possible directions for future work.

\subsection{Input Data Representation Methods}
Various factors are considered to determine input data representation methods. One is the entity type of a given task. For example, as a vehicle can only move on the road, using a grid representation may be inefficient owing to sparsity and overlap of the roads. In contrast, as humans can move freely, grid representation has been used in several studies. In STTP for humans, however, the input need not be represented in grid format for all cases. For instance, there could be a stronger relationship that is irrelevant to the distance or proximity (as mentioned in Section \ref{sec:rep}). In this case, locality would not be the most important information; thus grid representation may not be effective.

Recently, graph representation has been widely adopted in several applications. Any spatial unit, such as grid and road segment, can be represented as a node in a graph. As graphs can express arbitrary relationships between nodes, they may be a possible universal representation method for traffic networks with complex latent dependencies. However, to construct a graph representation, the adjacency matrix should be first obtained using domain knowledge. Thus, the graph represents only the pre-defined spatial dependency.

Finally, input data representation methods should be determined in consideration of various factors, including available dataset types, size of available datasets, domain knowledge, task characteristics, and deep network type. Thus, an appropriate design scheme for input data representation methods, such as analyzing which factor should be prioritized, and developing a new type of representation method, is required.

\subsection{DNN Techniques}
In most cases, each deep network employs a specific inductive bias. Through data analysis or domain knowledge, we can guess which inductive bias should be tried and enforced for each STTP problem. For instance, if there is a large dependency between proximate regions or timestamps, and this dependency is translation invariant, locality and translation invariance should be presented through the convolutional layers. By contrast, if there is no dependency between local regions or timestamps, convolutional layers would be less likely to be helpful. The availability and quality of dataset also plays an important role on deciding the DNN techniques to try. Even when the appropriate inductive bias is obvious for the given problem, we can employ a specific type of deep network only when the necessary dataset is available such that the information is extractable with deep networks. 

By the no-free-lunch theorem \cite{wolpert1997nolunch}, there is no consistently best-performing network for all the tasks. Therefore, it is inevitable to try multiple DNN models and choose the one with the best validation performance. When trying multiple models, knowledge on DNN technique's inductive bias and the characteristics of task and data should be fully utilized for an efficient modeling under limited time and computing resources. Furthermore, it can be helpful or even critical to consider criteria other than performance such as the required data size and training time.

\subsection{Dataset and Code Availability}
As mentioned in Section \ref{sec:dataset}, there is no clearly established standard datasets in the STTP area for benchmarking performance and integrating new findings. Therefore, it will be very helpful to agree on a few common datasets for each application area. Table \ref{table:dataset} can be a good start. This is an important matter especially for deep-learning based research. In deep-learning research, it is well known that the prediction accuracy for a specific dataset or task can be very sensitive to the choice of hyperparameters. For instance, modifying network depth or width can cause the performance ranking of the studied methods to be altered or even completely reversed. A sound study requires sufficient and fair amount of effort to be applied to each method such that meaningful findings can be reported. When each study uses its own dataset without other research groups verifying the results, it becomes difficult to tell if the findings are due to hyperparameter tuning or due to the newly proposed idea. Reproducibility is also an important issue in deep-learning research.  For this reason, ideally not only the dataset but also the code should be made available.

\subsection{Future Mobilities}
Recently, various new transportation systems, such as personal transporters (e.g., electric scooters, electric kickboards, and hoverboards), have been introduced. These vehicles are considerably smaller and more personalized, and their number is expected to continue to increase. Therefore, a proper management system as a part of the ITS should be developed to prevent accidents and congestion, and the relevant STTP algorithms should be studied.
These new mobilities have no movement constraints as in the case of humans, but they are significantly faster with more complex movement patterns. As new datasets are released, appropriate studies should be conducted. Similarly, autonomous vehicles might become a crucial component of transportation networks \cite{noruzoliaee2018roads}, and it might lead to an active development of STTP models for a mixed autonomous/human driving environment. 

Drone is another example of future mobilities. Until now, drones have been mainly used for freight transportation. In the near future, however, we can expect them to be used for passenger transportation as well. Drones have quite different characteristics compared to the traditional transportation systems. They move in airspace and can change their routes in three dimensional space. They need to consider geometry of the local obstacles such as houses and buildings. Their operation is significantly affected by the weather condition. Therefore, understanding the use of deep neural networks in the application area of drone will be an important and interesting topic.

\section{Acknowledgment}
This work was supported in part by a National Research Foundation of Korea (NRF) grant funded by the Korea government (MSIT) (No. NRF-2020R1A2C2007139) and Electronics  and  Telecommunications  Research  Institute  (ETRI)  grant  funded  by  the  Korean  government.  [20ZR1100,  Core Technologies of Distributed Intelligence Things  for Solving Industry and Society Problems]. 

\bibliographystyle{elsarticle-num}
\bibliography{references}

\begin{thebibliography}{100}
\expandafter\ifx\csname url\endcsname\relax
  \def\url#1{\texttt{#1}}\fi
\expandafter\ifx\csname urlprefix\endcsname\relax\def\urlprefix{URL }\fi
\expandafter\ifx\csname href\endcsname\relax
  \def\href#1#2{#2} \def\path#1{#1}\fi

\bibitem{veres2019survey}
M.~Veres, M.~Moussa, Deep learning for intelligent transportation systems: A
  survey of emerging trends, IEEE Transactions on Intelligent transportation
  systems (2019).

\bibitem{zhu2018itsbigdata}
L.~Zhu, F.~R. Yu, Y.~Wang, B.~Ning, T.~Tang, Big data analytics in intelligent
  transportation systems: A survey, IEEE Transactions on Intelligent
  Transportation Systems 20~(1) (2018) 383--398.

\bibitem{zhang2011itsdata}
J.~Zhang, F.-Y. Wang, K.~Wang, W.-H. Lin, X.~Xu, C.~Chen, Data-driven
  intelligent transportation systems: A survey, IEEE Transactions on
  Intelligent Transportation Systems 12~(4) (2011) 1624--1639.

\bibitem{touvron2020imagenet}
H.~Touvron, A.~Vedaldi, M.~Douze, H.~J{'e}gou, Fixing the train-test resolution
  discrepancy: Fixefficientnet, arXiv preprint arXiv:2003.08237 (2020).

\bibitem{xie2020imagenet2}
Q.~Xie, M.-T. Luong, E.~Hovy, Q.~V. Le, Self-training with noisy student
  improves imagenet classification, in: Proceedings of the IEEE/CVF Conference
  on Computer Vision and Pattern Recognition, 2020, pp. 10687--10698.

\bibitem{saon2017speech}
G.~Saon, G.~Kurata, T.~Sercu, K.~Audhkhasi, S.~Thomas, D.~Dimitriadis, X.~Cui,
  B.~Ramabhadran, M.~Picheny, L.-L. Lim, et~al., English conversational
  telephone speech recognition by humans and machines, arXiv preprint
  arXiv:1703.02136 (2017).

\bibitem{xiong2016speech2}
W.~Xiong, J.~Droppo, X.~Huang, F.~Seide, M.~Seltzer, A.~Stolcke, D.~Yu,
  G.~Zweig, Achieving human parity in conversational speech recognition, arXiv
  preprint arXiv:1610.05256 (2016).

\bibitem{zhang2016rethinking}
C.~Zhang, S.~Bengio, M.~Hardt, B.~Recht, O.~Vinyals, Understanding deep
  learning requires rethinking generalization, arXiv preprint arXiv:1611.03530
  (2016).

\bibitem{battaglia2018inductivebias}
P.~W. Battaglia, J.~B. Hamrick, V.~Bapst, A.~Sanchez-Gonzalez, V.~Zambaldi,
  M.~Malinowski, A.~Tacchetti, D.~Raposo, A.~Santoro, R.~Faulkner, et~al.,
  Relational inductive biases, deep learning, and graph networks, arXiv
  preprint arXiv:1806.01261 (2018).

\bibitem{tobler1970computer}
W.~R. Tobler, A computer movie simulating urban growth in the detroit region,
  Economic geography 46~(sup1) (1970) 234--240.

\bibitem{cifar10dataset}
A.~Krizhevsky, V.~Nair, G.~Hinton,
  \href{http://www.cs.toronto.edu/~kriz/cifar.html}{Cifar-10 (canadian
  institute for advanced research)}.
\newline\urlprefix\url{http://www.cs.toronto.edu/~kriz/cifar.html}

\bibitem{cifar100dataset}
.~Krizhevsky, Cifar100.

\bibitem{imagenet_cvpr09}
J.~Deng, W.~Dong, R.~Socher, L.-J. Li, K.~Li, L.~Fei-Fei, {ImageNet: A
  Large-Scale Hierarchical Image Database}, in: CVPR09, 2009.

\bibitem{liu2017crowdflow}
L.~Liu, R.-C. Chen, A novel passenger flow prediction model using deep learning
  methods, Transportation Research Part C: Emerging Technologies 84 (2017)
  74--91.

\bibitem{li2017trafficspeed}
Y.~Li, R.~Yu, C.~Shahabi, Y.~Liu, Diffusion convolutional recurrent neural
  network: Data-driven traffic forecasting, arXiv preprint arXiv:1707.01926
  (2017).

\bibitem{yu2017trafficspeed}
R.~Yu, Y.~Li, C.~Shahabi, U.~Demiryurek, Y.~Liu, Deep learning: A generic
  approach for extreme condition traffic forecasting, in: Proceedings of the
  2017 SIAM international Conference on Data Mining, SIAM, 2017, pp. 777--785.

\bibitem{chu2018crowddemand}
J.~Chu, K.~Qian, X.~Wang, L.~Yao, F.~Xiao, J.~Li, X.~Miao, Z.~Yang, Passenger
  demand prediction with cellular footprints, in: 2018 15th Annual IEEE
  International Conference on Sensing, Communication, and Networking (SECON),
  IEEE, 2018, pp. 1--9.

\bibitem{lin2018crowddemand}
L.~Lin, Z.~He, S.~Peeta, Predicting station-level hourly demand in a
  large-scale bike-sharing network: A graph convolutional neural network
  approach, Transportation Research Part C: Emerging Technologies 97 (2018)
  258--276.

\bibitem{chai2018crowdflow}
D.~Chai, L.~Wang, Q.~Yang, Bike flow prediction with multi-graph convolutional
  networks, in: Proceedings of the 26th ACM SIGSPATIAL International Conference
  on Advances in Geographic Information Systems, 2018, pp. 397--400.

\bibitem{cui2019trafficspeed}
Z.~Cui, K.~Henrickson, R.~Ke, Y.~Wang, Traffic graph convolutional recurrent
  neural network: A deep learning framework for network-scale traffic learning
  and forecasting, IEEE Transactions on Intelligent Transportation Systems
  (2019).

\bibitem{iyer2018trafficspeed}
S.~R. Iyer, K.~Boxer, L.~Subramanian, Urban traffic congestion mapping using
  bus mobility data., in: UMCit@ KDD, 2018, pp. 7--13.

\bibitem{lv2018trafficspeed}
Z.~Lv, J.~Xu, K.~Zheng, H.~Yin, P.~Zhao, X.~Zhou, Lc-rnn: A deep learning model
  for traffic speed prediction., in: IJCAI, 2018, pp. 3470--3476.

\bibitem{zhao2019trafficspeed}
J.~Zhao, Y.~Gao, Z.~Yang, J.~Li, Y.~Feng, Z.~Qin, Z.~Bai, Truck traffic speed
  prediction under non-recurrent congestion: Based on optimized deep learning
  algorithms and gps data, IEEE Access 7 (2019) 9116--9127.

\bibitem{zhang2019trafficspeed}
Z.~Zhang, M.~Li, X.~Lin, Y.~Wang, F.~He, Multistep speed prediction on traffic
  networks: A deep learning approach considering spatio-temporal dependencies,
  Transportation research part C: emerging technologies 105 (2019) 297--322.

\bibitem{zhang2019trafficspeedflow}
Y.~Zhang, T.~Cheng, Y.~Ren, K.~Xie, A novel residual graph convolution deep
  learning model for short-term network-based traffic forecasting,
  International Journal of Geographical Information Science (2019) 1--27.

\bibitem{du2020crowddemand}
B.~Du, X.~Hu, L.~Sun, J.~Liu, Y.~Qiao, W.~Lv, Traffic demand prediction based
  on dynamic transition convolutional neural network, IEEE Transactions on
  Intelligent Transportation Systems (2020).

\bibitem{lv2020trafficflow}
M.~Lv, Z.~Hong, L.~Chen, T.~Chen, T.~Zhu, S.~Ji, Temporal multi-graph
  convolutional network for traffic flow prediction, IEEE Transactions on
  Intelligent Transportation Systems (2020).

\bibitem{lee2020trafficspeed}
K.~Lee, W.~Rhee, \href{http://arxiv.org/abs/1905.12256}{Graph convolutional
  modules for traffic forecasting}, CoRR abs/1905.12256 (2019).
\newblock \href {http://arxiv.org/abs/1905.12256} {\path{arXiv:1905.12256}}.
\newline\urlprefix\url{http://arxiv.org/abs/1905.12256}

\bibitem{ye2020graphsurvey}
J.~Ye, J.~Zhao, K.~Ye, C.~Xu, How to build a graph-based deep learning
  architecture in traffic domain: A survey, arXiv preprint arXiv:2005.11691
  (2020).

\bibitem{zhang2017crowdflow}
J.~Zhang, Y.~Zheng, D.~Qi, Deep spatio-temporal residual networks for citywide
  crowd flows prediction, in: Thirty-First AAAI Conference on Artificial
  Intelligence, 2017.

\bibitem{earlysuccess}
A.~Krizhevsky, I.~Sutskever, G.~E. Hinton,
  \href{http://papers.nips.cc/paper/4824-imagenet-classification-with-deep-convolutional-neural-networks.pdf}{Imagenet
  classification with deep convolutional neural networks}, in: F.~Pereira,
  C.~J.~C. Burges, L.~Bottou, K.~Q. Weinberger (Eds.), Advances in Neural
  Information Processing Systems 25, Curran Associates, Inc., 2012, pp.
  1097--1105.
\newline\urlprefix\url{http://papers.nips.cc/paper/4824-imagenet-classification-with-deep-convolutional-neural-networks.pdf}

\bibitem{firstmlp}
K.~Hornik, M.~Stinchcombe, H.~White, Multilayer feedforward networks are
  universal approximators, Neural Netw. 2~(5) (1989) 359–366.

\bibitem{firstcnnjapan}
K.~Fukushima, S.~Miyake, Neocognitron: A self-organizing neural network model
  for a mechanism of visual pattern recognition, in: Competition and
  cooperation in neural nets, Springer, 1982, pp. 267--285.

\bibitem{firstcnnlecun}
Y.~LeCun, P.~Haffner, L.~Bottou, Y.~Bengio, Object recognition with
  gradient-based learning, in: Shape, contour and grouping in computer vision,
  Springer, 1999, pp. 319--345.

\bibitem{firstrnn}
M.~I. Jordan, Serial order: A parallel distributed processing approach, in:
  Advances in psychology, Vol. 121, Elsevier, 1997, pp. 471--495.

\bibitem{pcahinton}
G.~E. Hinton, R.~R. Salakhutdinov,
  \href{http://www.ncbi.nlm.nih.gov/sites/entrez?db=pubmed&uid=16873662&cmd=showdetailview&indexed=google}{Reducing
  the dimensionality of data with neural networks}, Science 313~(5786) (2006)
  504--507.
\newblock \href {https://doi.org/10.1126/science.1127647}
  {\path{doi:10.1126/science.1127647}}.
\newline\urlprefix\url{http://www.ncbi.nlm.nih.gov/sites/entrez?db=pubmed&uid=16873662&cmd=showdetailview&indexed=google}

\bibitem{bai2018empirical}
S.~Bai, J.~Z. Kolter, V.~Koltun, An empirical evaluation of generic
  convolutional and recurrent networks for sequence modeling, arXiv preprint
  arXiv:1803.01271 (2018).

\bibitem{tang2018self}
G.~Tang, M.~M{\"u}ller, A.~Rios, R.~Sennrich, Why self-attention? a targeted
  evaluation of neural machine translation architectures, arXiv preprint
  arXiv:1808.08946 (2018).

\bibitem{firstperceptron}
F.~Rosenblatt, Principles of neurodynamics. perceptrons and the theory of brain
  mechanisms, Tech. rep., Cornell Aeronautical Lab Inc Buffalo NY (1961).

\bibitem{goodfellow2016deep}
I.~Goodfellow, Y.~Bengio, A.~Courville, Y.~Bengio, Deep learning, Vol.~1, MIT
  press Cambridge, 2016.

\bibitem{wang2016trafficspeed}
J.~Wang, Q.~Gu, J.~Wu, G.~Liu, Z.~Xiong, Traffic speed prediction and
  congestion source exploration: A deep learning method, in: 2016 IEEE 16th
  International Conference on Data Mining (ICDM), IEEE, 2016, pp. 499--508.

\bibitem{ma2017trafficspeed}
X.~Ma, Z.~Dai, Z.~He, J.~Ma, Y.~Wang, Y.~Wang, Learning traffic as images: a
  deep convolutional neural network for large-scale transportation network
  speed prediction, Sensors 17~(4) (2017) 818.

\bibitem{shen2018trafficspeed}
G.~Shen, C.~Chen, Q.~Pan, S.~Shen, Z.~Liu, Research on traffic speed prediction
  by temporal clustering analysis and convolutional neural network with
  deformable kernels (may, 2018), IEEE Access 6 (2018) 51756--51765.

\bibitem{rabiner1989tutorial}
L.~R. Rabiner, A tutorial on hidden markov models and selected applications in
  speech recognition, Proceedings of the IEEE 77~(2) (1989) 257--286.

\bibitem{lstm}
S.~Hochreiter, J.~Schmidhuber, Long short-term memory, Neural computation 9~(8)
  (1997) 1735--1780.

\bibitem{vinyals2015show}
O.~Vinyals, A.~Toshev, S.~Bengio, D.~Erhan, Show and tell: A neural image
  caption generator, in: Proceedings of the IEEE conference on computer vision
  and pattern recognition, 2015, pp. 3156--3164.

\bibitem{wu2020comprehensive}
Z.~Wu, S.~Pan, F.~Chen, G.~Long, C.~Zhang, S.~Y. Philip, A comprehensive survey
  on graph neural networks, IEEE Transactions on Neural Networks and Learning
  Systems (2020).

\bibitem{bruna2013spectral}
J.~Bruna, W.~Zaremba, A.~Szlam, Y.~LeCun, Spectral networks and locally
  connected networks on graphs, arXiv preprint arXiv:1312.6203 (2013).

\bibitem{pan2009survey}
S.~J. Pan, Q.~Yang, A survey on transfer learning, IEEE Transactions on
  knowledge and data engineering 22~(10) (2009) 1345--1359.

\bibitem{lake2015human}
B.~M. Lake, R.~Salakhutdinov, J.~B. Tenenbaum, Human-level concept learning
  through probabilistic program induction, Science 350~(6266) (2015)
  1332--1338.

\bibitem{carey1978acquiring}
S.~Carey, E.~Bartlett, Acquiring a single new word. (1978).

\bibitem{lake2015omniglot}
B.~M. Lake, R.~Salakhutdinov, J.~B. Tenenbaum, Human-level concept learning
  through probabilistic program induction, Science 350~(6266) (2015)
  1332--1338.

\bibitem{silver2016mastering}
D.~Silver, A.~Huang, C.~J. Maddison, A.~Guez, L.~Sifre, G.~Van Den~Driessche,
  J.~Schrittwieser, I.~Antonoglou, V.~Panneershelvam, M.~Lanctot, et~al.,
  Mastering the game of go with deep neural networks and tree search, nature
  529~(7587) (2016) 484--489.

\bibitem{silver2017mastering}
D.~Silver, J.~Schrittwieser, K.~Simonyan, I.~Antonoglou, A.~Huang, A.~Guez,
  T.~Hubert, L.~Baker, M.~Lai, A.~Bolton, et~al., Mastering the game of go
  without human knowledge, nature 550~(7676) (2017) 354--359.

\bibitem{vaswani2017attention}
A.~Vaswani, N.~Shazeer, N.~Parmar, J.~Uszkoreit, L.~Jones, A.~N. Gomez,
  {\L}.~Kaiser, I.~Polosukhin, Attention is all you need, in: Advances in
  neural information processing systems, 2017, pp. 5998--6008.

\bibitem{cho2014learning}
K.~Cho, B.~Van~Merri{\"e}nboer, C.~Gulcehre, D.~Bahdanau, F.~Bougares,
  H.~Schwenk, Y.~Bengio, Learning phrase representations using rnn
  encoder-decoder for statistical machine translation, arXiv preprint
  arXiv:1406.1078 (2014).

\bibitem{devlin2018bert}
J.~Devlin, M.-W. Chang, K.~Lee, K.~Toutanova, Bert: Pre-training of deep
  bidirectional transformers for language understanding, arXiv preprint
  arXiv:1810.04805 (2018).

\bibitem{kingma2013auto}
D.~P. Kingma, M.~Welling, Auto-encoding variational bayes, arXiv preprint
  arXiv:1312.6114 (2013).

\bibitem{pu2016variational}
Y.~Pu, Z.~Gan, R.~Henao, X.~Yuan, C.~Li, A.~Stevens, L.~Carin, Variational
  autoencoder for deep learning of images, labels and captions, in: Advances in
  neural information processing systems, 2016, pp. 2352--2360.

\bibitem{goodfellow2014generative}
I.~Goodfellow, J.~Pouget-Abadie, M.~Mirza, B.~Xu, D.~Warde-Farley, S.~Ozair,
  A.~Courville, Y.~Bengio, Generative adversarial nets, in: Advances in neural
  information processing systems, 2014, pp. 2672--2680.

\bibitem{lan2019albert}
Z.~Lan, M.~Chen, S.~Goodman, K.~Gimpel, P.~Sharma, R.~Soricut, Albert: A lite
  bert for self-supervised learning of language representations, arXiv preprint
  arXiv:1909.11942 (2019).

\bibitem{chen2020simple}
T.~Chen, S.~Kornblith, M.~Norouzi, G.~Hinton, A simple framework for
  contrastive learning of visual representations, arXiv preprint
  arXiv:2002.05709 (2020).

\bibitem{chen2018crowdflow}
C.~Chen, K.~Li, S.~G. Teo, G.~Chen, X.~Zou, X.~Yang, R.~C. Vijay, J.~Feng,
  Z.~Zeng, Exploiting spatio-temporal correlations with multiple 3d
  convolutional neural networks for citywide vehicle flow prediction, in: 2018
  IEEE international conference on data mining (ICDM), IEEE, 2018, pp.
  893--898.

\bibitem{wang2018crowdflow}
B.~Wang, Z.~Yan, J.~Lu, G.~Zhang, T.~Li, Explore uncertainty in residual
  networks for crowds flow prediction, in: 2018 International Joint Conference
  on Neural Networks (IJCNN), IEEE, 2018, pp. 1--7.

\bibitem{zhang2018crowdflowcnn}
J.~Zhang, Y.~Zheng, D.~Qi, R.~Li, X.~Yi, T.~Li, Predicting citywide crowd flows
  using deep spatio-temporal residual networks, Artificial Intelligence 259
  (2018) 147--166.

\bibitem{lin2019crowdflow}
Z.~Lin, J.~Feng, Z.~Lu, Y.~Li, D.~Jin, Deepstn+: Context-aware spatial-temporal
  neural network for crowd flow prediction in metropolis, in: Proceedings of
  the AAAI Conference on Artificial Intelligence, Vol.~33, 2019, pp.
  1020--1027.

\bibitem{rong2019crowdflow}
C.~Rong, J.~Feng, Y.~Li, Deep learning models for population flow generation
  from aggregated mobility data, in: Adjunct Proceedings of the 2019 ACM
  International Joint Conference on Pervasive and Ubiquitous Computing and
  Proceedings of the 2019 ACM International Symposium on Wearable Computers,
  2019, pp. 1008--1013.

\bibitem{zhang2019crowdflow}
J.~Zhang, Y.~Zheng, J.~Sun, D.~Qi, Flow prediction in spatio-temporal networks
  based on multitask deep learning, IEEE Transactions on Knowledge and Data
  Engineering (2019).

\bibitem{ma2018crowdflow}
X.~Ma, J.~Zhang, B.~Du, C.~Ding, L.~Sun, Parallel architecture of convolutional
  bi-directional lstm neural networks for network-wide metro ridership
  prediction, IEEE Transactions on Intelligent Transportation Systems 20~(6)
  (2018) 2278--2288.

\bibitem{zonoozi2018crowdflow}
A.~Zonoozi, J.-j. Kim, X.-L. Li, G.~Cong, Periodic-crn: A convolutional
  recurrent model for crowd density prediction with recurring periodic
  patterns., in: IJCAI, 2018, pp. 3732--3738.

\bibitem{ren2019crowdflow}
Y.~Ren, H.~Chen, Y.~Han, T.~Cheng, Y.~Zhang, G.~Chen, A hybrid integrated deep
  learning model for the prediction of citywide spatio-temporal flow volumes,
  International Journal of Geographical Information Science (2019) 1--22.

\bibitem{li2019crowdflow}
W.~Li, W.~Tao, J.~Qiu, X.~Liu, X.~Zhou, Z.~Pan, Densely connected convolutional
  networks with attention lstm for crowd flows prediction, IEEE Access 7 (2019)
  140488--140498.

\bibitem{huang2017crowddemand}
Z.~Huang, Z.~Zhao, E.~Shijia, C.~Yu, G.~Shan, T.~Li, J.~Cheng, J.~Sun,
  Y.~Xiang, Prace: A taxi recommender for finding passengers with deep learning
  approaches, in: International Conference on Intelligent Computing, Springer,
  2017, pp. 759--770.

\bibitem{ke2017crowddemand}
J.~Ke, H.~Zheng, H.~Yang, X.~M. Chen, Short-term forecasting of passenger
  demand under on-demand ride services: A spatio-temporal deep learning
  approach, Transportation Research Part C: Emerging Technologies 85 (2017)
  591--608.

\bibitem{liao2018crowddemand}
S.~Liao, L.~Zhou, X.~Di, B.~Yuan, J.~Xiong, Large-scale short-term urban taxi
  demand forecasting using deep learning, in: 2018 23rd Asia and South Pacific
  Design Automation Conference (ASP-DAC), IEEE, 2018, pp. 428--433.

\bibitem{wang2018crowddemand}
D.~Wang, Y.~Yang, S.~Ning, Deepstcl: A deep spatio-temporal convlstm for travel
  demand prediction, in: 2018 International Joint Conference on Neural Networks
  (IJCNN), IEEE, 2018, pp. 1--8.

\bibitem{yao2018crowddemand}
H.~Yao, F.~Wu, J.~Ke, X.~Tang, Y.~Jia, S.~Lu, P.~Gong, J.~Ye, Z.~Li, Deep
  multi-view spatial-temporal network for taxi demand prediction, in:
  Thirty-Second AAAI Conference on Artificial Intelligence, 2018.

\bibitem{zhou2018crowdflow}
X.~Zhou, Y.~Shen, Y.~Zhu, L.~Huang, Predicting multi-step citywide passenger
  demands using attention-based neural networks, in: Proceedings of the
  Eleventh ACM International Conference on Web Search and Data Mining, 2018,
  pp. 736--744.

\bibitem{yao2019crowddemand}
H.~Yao, Y.~Liu, Y.~Wei, X.~Tang, Z.~Li, Learning from multiple cities: A
  meta-learning approach for spatial-temporal prediction, in: The World Wide
  Web Conference, 2019, pp. 2181--2191.

\bibitem{he2019crowddemand}
S.~He, K.~G. Shin, Spatio-temporal capsule-based reinforcement learning for
  mobility-on-demand network coordination, in: The World Wide Web Conference,
  2019, pp. 2806--2813.

\bibitem{zhou2020crowddemand}
Y.~Zhou, J.~Li, H.~Chen, Y.~Wu, J.~Wu, L.~Chen, A spatiotemporal attention
  mechanism-based model for multi-step citywide passenger demand prediction,
  Information Sciences 513 (2020) 372--385.

\bibitem{daganzo2019general}
C.~F. Daganzo, Y.~Ouyang, A general model of demand-responsive transportation
  services: From taxi to ridesharing to dial-a-ride, Transportation Research
  Part B: Methodological 126 (2019) 213--224.

\bibitem{geng2019crowddemand}
X.~Geng, Y.~Li, L.~Wang, L.~Zhang, Q.~Yang, J.~Ye, Y.~Liu, Spatiotemporal
  multi-graph convolution network for ride-hailing demand forecasting, in:
  Proceedings of the AAAI Conference on Artificial Intelligence, Vol.~33, 2019,
  pp. 3656--3663.

\bibitem{lv2014trafficflow}
Y.~Lv, Y.~Duan, W.~Kang, Z.~Li, F.-Y. Wang, Traffic flow prediction with big
  data: a deep learning approach, IEEE Transactions on Intelligent
  Transportation Systems 16~(2) (2014) 865--873.

\bibitem{baek2016crowdflow}
J.~Baek~and, K.~Sohn, Deep-learning architectures to forecast bus ridership at
  the stop and stop-to-stop levels for dense and crowded bus networks, Applied
  Artificial Intelligence 30~(9) (2016) 861--885.

\bibitem{duan2016trafficflow}
Y.~Duan, Y.~Lv, Y.-L. Liu, F.-Y. Wang, An efficient realization of deep
  learning for traffic data imputation, Transportation research part C:
  emerging technologies 72 (2016) 168--181.

\bibitem{koesdwiady2016trafficflow}
A.~Koesdwiady, R.~Soua, F.~Karray, Improving traffic flow prediction with
  weather information in connected cars: A deep learning approach, IEEE
  Transactions on Vehicular Technology 65~(12) (2016) 9508--9517.

\bibitem{yang2016trafficflow}
H.-F. Yang, T.~S. Dillon, Y.-P.~P. Chen, Optimized structure of the traffic
  flow forecasting model with a deep learning approach, IEEE transactions on
  neural networks and learning systems 28~(10) (2016) 2371--2381.

\bibitem{jia2016trafficspeed}
Y.~Jia, J.~Wu, Y.~Du, Traffic speed prediction using deep learning method, in:
  2016 IEEE 19th International Conference on Intelligent Transportation Systems
  (ITSC), IEEE, 2016, pp. 1217--1222.

\bibitem{jia2017trafficflow}
Y.~Jia, J.~Wu, M.~Xu, Traffic flow prediction with rainfall impact using a deep
  learning method, Journal of advanced transportation 2017 (2017).

\bibitem{pamula2018trafficflow}
T.~Pamu{\l}a, Impact of data loss for prediction of traffic flow on an urban
  road using neural networks, IEEE Transactions on Intelligent Transportation
  Systems 20~(3) (2018) 1000--1009.

\bibitem{zhang2018trafficflow}
Y.~Zhang, G.~Huang, Traffic flow prediction model based on deep belief network
  and genetic algorithm, IET Intelligent Transport Systems 12~(6) (2018)
  533--541.

\bibitem{tian2015trafficflow}
Y.~Tian, L.~Pan, Predicting short-term traffic flow by long short-term memory
  recurrent neural network, in: 2015 IEEE international conference on smart
  city/SocialCom/SustainCom (SmartCity), IEEE, 2015, pp. 153--158.

\bibitem{ma2015trafficspeed}
X.~Ma, Z.~Tao, Y.~Wang, H.~Yu, Y.~Wang, Long short-term memory neural network
  for traffic speed prediction using remote microwave sensor data,
  Transportation Research Part C: Emerging Technologies 54 (2015) 187--197.

\bibitem{fu2016trafficflow}
R.~Fu, Z.~Zhang, L.~Li, Using lstm and gru neural network methods for traffic
  flow prediction, in: 2016 31st Youth Academic Annual Conference of Chinese
  Association of Automation (YAC), IEEE, 2016, pp. 324--328.

\bibitem{elhenawy2016trafficflow}
M.~Elhenawy, H.~Rakha, Stretch-wide traffic state prediction using
  discriminatively pre-trained deep neural networks, in: 2016 IEEE 19th
  International Conference on Intelligent Transportation Systems (ITSC), IEEE,
  2016, pp. 1065--1070.

\bibitem{wang2017crowddemand}
D.~Wang, W.~Cao, J.~Li, J.~Ye, Deepsd: Supply-demand prediction for online
  car-hailing services using deep neural networks, in: 2017 IEEE 33rd
  international conference on data engineering (ICDE), IEEE, 2017, pp.
  243--254.

\bibitem{sudo2017crowdflow}
A.~Sudo, T.-H. Teng, H.~C. Lau, Y.~Sekimoto, Predicting indoor crowd density
  using column-structured deep neural network, in: Proceedings of the 1st ACM
  SIGSPATIAL Workshop on Prediction of Human Mobility, 2017, pp. 1--7.

\bibitem{yuan2017trafficaccidents}
Z.~Yuan, X.~Zhou, T.~Yang, J.~Tamerius, R.~Mantilla, Predicting traffic
  accidents through heterogeneous urban data: A case study, in: Proceedings of
  the 6th International Workshop on Urban Computing (UrbComp 2017), Halifax,
  NS, Canada, Vol.~14, 2017.

\bibitem{dai2017trafficflow}
X.~Dai, R.~Fu, Y.~Lin, L.~Li, F.-Y. Wang, Deeptrend: A deep hierarchical neural
  network for traffic flow prediction, arXiv preprint arXiv:1707.03213 (2017).

\bibitem{kang2017trafficflow}
D.~Kang, Y.~Lv, Y.-y. Chen, Short-term traffic flow prediction with lstm
  recurrent neural network, in: 2017 IEEE 20th International Conference on
  Intelligent Transportation Systems (ITSC), IEEE, 2017, pp. 1--6.

\bibitem{zhao2017trafficflow}
Z.~Zhao, W.~Chen, X.~Wu, P.~C. Chen, J.~Liu, Lstm network: a deep learning
  approach for short-term traffic forecast, IET Intelligent Transport Systems
  11~(2) (2017) 68--75.

\bibitem{jia2017trafficspeed}
Y.~Jia, J.~Wu, M.~Ben-Akiva, R.~Seshadri, Y.~Du, Rainfall-integrated traffic
  speed prediction using deep learning method, IET Intelligent Transport
  Systems 11~(9) (2017) 531--536.

\bibitem{polson2017trafficspeed}
N.~G. Polson, V.~O. Sokolov, Deep learning for short-term traffic flow
  prediction, Transportation Research Part C: Emerging Technologies 79 (2017)
  1--17.

\bibitem{sun2017trafficcongestion}
F.~Sun, A.~Dubey, J.~White, Dxnat - deep neural networks for explaining
  non-recurring traffic congestion, in: 2017 IEEE International Conference on
  Big Data (Big Data), IEEE, 2017, pp. 2141--2150.

\bibitem{xu2018crowddemand}
C.~Xu, J.~Ji, P.~Liu, The station-free sharing bike demand forecasting with a
  deep learning approach and large-scale datasets, Transportation research part
  C: emerging technologies 95 (2018) 47--60.

\bibitem{abbas2018trafficflow}
Z.~Abbas, A.~Al-Shishtawy, S.~Girdzijauskas, V.~Vlassov, Short-term traffic
  prediction using long short-term memory neural networks, in: 2018 IEEE
  International Congress on Big Data (BigData Congress), IEEE, 2018, pp.
  57--65.

\bibitem{xie2018trafficflow}
Z.~Xie, Q.~Liu, Lstm networks for vessel traffic flow prediction in inland
  waterway, in: 2018 IEEE International Conference on Big Data and Smart
  Computing (BigComp), IEEE, 2018, pp. 418--425.

\bibitem{zhang2018trafficflowrnn}
D.~Zhang, M.~R. Kabuka, Combining weather condition data to predict traffic
  flow: a gru-based deep learning approach, IET Intelligent Transport Systems
  12~(7) (2018) 578--585.

\bibitem{zhao2018trafficflow}
B.~Zhao, X.~Zhang, A parallel-res gru architecture and its application to road
  network traffic flow forecasting, in: Proceedings of 2018 International
  Conference on Big Data Technologies, 2018, pp. 79--83.

\bibitem{adu2019trafficspeed}
Y.~Adu-Gyamfi, M.~Zhao, Traffic speed prediction for urban arterial roads using
  deep neural networks, Journal of Computing in Civil Engineering (2019).

\bibitem{cui2018trafficspeedrnn}
Z.~Cui, R.~Ke, Z.~Pu, Y.~Wang, Deep bidirectional and unidirectional lstm
  recurrent neural network for network-wide traffic speed prediction, arXiv
  preprint arXiv:1801.02143 (2018).

\bibitem{fandango2018trafficspeed}
A.~Fandango, R.~P. Wiegand, Towards investigation of iterative strategy for
  data mining of short-term traffic flow with recurrent neural networks, in:
  Proceedings of the 2nd International Conference on Information System and
  Data Mining, 2018, pp. 65--69.

\bibitem{liao2018trafficspeed}
B.~Liao, J.~Zhang, M.~Cai, S.~Tang, Y.~Gao, C.~Wu, S.~Yang, W.~Zhu, Y.~Guo,
  F.~Wu, Dest-resnet: A deep spatiotemporal residual network for hotspot
  traffic speed prediction, in: Proceedings of the 26th ACM international
  conference on Multimedia, 2018, pp. 1883--1891.

\bibitem{xiangxue2019trafficflow}
W.~Xiangxue, X.~Lunhui, C.~Kaixun, Data-driven short-term forecasting for urban
  road network traffic based on data processing and lstm-rnn, Arabian Journal
  for Science and Engineering 44~(4) (2019) 3043--3060.

\bibitem{sun2019trafficcongestion}
S.~Sun, J.~Chen, J.~Sun, Traffic congestion prediction based on gps trajectory
  data, International Journal of Distributed Sensor Networks 15~(5) (2019)
  1550147719847440.

\bibitem{wu2016trafficflow}
Y.~Wu, H.~Tan, Short-term traffic flow forecasting with spatial-temporal
  correlation in a hybrid deep learning framework, arXiv preprint
  arXiv:1612.01022 (2016).

\bibitem{du2017trafficflow}
S.~Du, T.~Li, X.~Gong, Y.~Yang, S.~J. Horng, Traffic flow forecasting based on
  hybrid deep learning framework, in: 2017 12th International Conference on
  Intelligent Systems and Knowledge Engineering (ISKE), IEEE, 2017, pp. 1--6.

\bibitem{liu2017trafficflow}
Y.~Liu, H.~Zheng, X.~Feng, Z.~Chen, Short-term traffic flow prediction with
  conv-lstm, in: 2017 9th International Conference on Wireless Communications
  and Signal Processing (WCSP), IEEE, 2017, pp. 1--6.

\bibitem{yu2017trafficspeedhybrid}
H.~Yu, Z.~Wu, S.~Wang, Y.~Wang, X.~Ma, Spatiotemporal recurrent convolutional
  networks for traffic prediction in transportation networks, Sensors 17~(7)
  (2017) 1501.

\bibitem{yang2018trafficflow}
G.~Yang, Y.~Wang, H.~Yu, Y.~Ren, J.~Xie, Short-term traffic state prediction
  based on the spatiotemporal features of critical road sections, Sensors
  18~(7) (2018) 2287.

\bibitem{duan2018trafficflow}
Z.~Duan, Y.~Yang, K.~Zhang, Y.~Ni, S.~Bajgain, Improved deep hybrid networks
  for urban traffic flow prediction using trajectory data, IEEE Access 6 (2018)
  31820--31827.

\bibitem{yu2017trafficspeedgraph}
B.~Yu, H.~Yin, Z.~Zhu, Spatio-temporal graph convolutional networks: A deep
  learning framework for traffic forecasting, arXiv preprint arXiv:1709.04875
  (2017).

\bibitem{cheng2018trafficflow}
X.~Cheng, R.~Zhang, J.~Zhou, W.~Xu, Deeptransport: Learning spatial-temporal
  dependency for traffic condition forecasting, in: 2018 International Joint
  Conference on Neural Networks (IJCNN), IEEE, 2018, pp. 1--8.

\bibitem{ning2018crowddemand}
X.~Ning, L.~Yao, X.~Wang, B.~Benatallah, F.~Salim, P.~D. Haghighi, Predicting
  citywide passenger demand via reinforcement learning from spatio-temporal
  dynamics, in: Proceedings of the 15th EAI International Conference on Mobile
  and Ubiquitous Systems: Computing, Networking and Services, 2018, pp. 19--28.

\bibitem{rodrigues2019crowddemand}
F.~Rodrigues, I.~Markou, F.~C. Pereira, Combining time-series and textual data
  for taxi demand prediction in event areas: A deep learning approach,
  Information Fusion 49 (2019) 120--129.

\bibitem{li2018crowddemand}
Y.~Li, Y.~Zheng, Q.~Yang, Dynamic bike reposition: A spatio-temporal
  reinforcement learning approach, in: Proceedings of the 24th ACM SIGKDD
  International Conference on Knowledge Discovery \& Data Mining, 2018, pp.
  1724--1733.

\bibitem{wang2018crowdflowtransfer}
L.~Wang, X.~Geng, X.~Ma, F.~Liu, Q.~Yang, Cross-city transfer learning for deep
  spatio-temporal prediction, arXiv preprint arXiv:1802.00386 (2018).

\bibitem{yao2018crowddemandatt}
H.~Yao, X.~Tang, H.~Wei, G.~Zheng, Y.~Yu, Z.~Li, Modeling spatial-temporal
  dynamics for traffic prediction, arXiv preprint arXiv:1803.01254 (2018).

\bibitem{du2018trafficflow}
S.~Du, T.~Li, X.~Gong, Z.~Yu, Y.~Huang, S.-J. Horng, A hybrid method for
  traffic flow forecasting using multimodal deep learning, arXiv preprint
  arXiv:1803.02099 (2018).

\bibitem{liang2018trafficflow}
Y.~Liang, Z.~Cui, Y.~Tian, H.~Chen, Y.~Wang, A deep generative adversarial
  architecture for network-wide spatial-temporal traffic-state estimation,
  Transportation Research Record 2672~(45) (2018) 87--105.

\bibitem{wu2018trafficflowgraphatt}
T.~Wu, F.~Chen, Y.~Wan, Graph attention lstm network: A new model for traffic
  flow forecasting, in: 2018 5th International Conference on Information
  Science and Control Engineering (ICISCE), IEEE, 2018, pp. 241--245.

\bibitem{wu2018trafficflowatt}
Y.~Wu, H.~Tan, L.~Qin, B.~Ran, Z.~Jiang, A hybrid deep learning based traffic
  flow prediction method and its understanding, Transportation Research Part C:
  Emerging Technologies 90 (2018) 166--180.

\bibitem{liu2018trafficspeed}
Q.~Liu, B.~Wang, Y.~Zhu, Short-term traffic speed forecasting based on
  attention convolutional neural network for arterials, Computer-Aided Civil
  and Infrastructure Engineering 33~(11) (2018) 999--1016.

\bibitem{ma2018trafficspeed}
X.~Ma, Y.~Li, Z.~Cui, Y.~Wang, Forecasting transportation network speed using
  deep capsule networks with nested lstm models, arXiv preprint
  arXiv:1811.04745 (2018).

\bibitem{he2018trafficspeed}
Z.~He, C.-Y. Chow, J.-D. Zhang, Stann: A spatio--temporal attentive neural
  network for traffic prediction, IEEE Access 7 (2018) 4795--4806.

\bibitem{mourad2019crowdflow}
L.~Mourad, H.~Qi, Y.~Shen, B.~Yin, Astir: Spatio-temporal data mining for crowd
  flow prediction, IEEE Access 7 (2019) 175159--175165.

\bibitem{zhou2019crowddemand}
Y.~Zhou, H.~Chen, J.~Li, Y.~Wu, J.~Wu, L.~Chen, St-attn: Spatial-temporal
  attention mechanism for multi-step citywide crowd flow prediction, in: 2019
  International Conference on Data Mining Workshops (ICDMW), IEEE, 2019, pp.
  609--614.

\bibitem{zhang2019trafficaccident}
Y.~Zhang, H.~Wang, D.~Zhang, D.~Wang, Deeprisk: A deep transfer learning
  approach to migratable traffic risk estimation in intelligent transportation
  using social sensing, in: 2019 15th International Conference on Distributed
  Computing in Sensor Systems (DCOSS), IEEE, 2019, pp. 123--130.

\bibitem{guo2019trafficflow}
S.~Guo, Y.~Lin, N.~Feng, C.~Song, H.~Wan, Attention based spatial-temporal
  graph convolutional networks for traffic flow forecasting, in: Proceedings of
  the AAAI Conference on Artificial Intelligence, Vol.~33, 2019, pp. 922--929.

\bibitem{pan2019trafficflow}
Z.~Pan, Y.~Liang, W.~Wang, Y.~Yu, Y.~Zheng, J.~Zhang, Urban traffic prediction
  from spatio-temporal data using deep meta learning, in: Proceedings of the
  25th ACM SIGKDD International Conference on Knowledge Discovery \& Data
  Mining, 2019, pp. 1720--1730.

\bibitem{zheng2019trafficspeed}
C.~Zheng, X.~Fan, C.~Wang, J.~Qi, Gman: A graph multi-attention network for
  traffic prediction, arXiv preprint arXiv:1911.08415 (2019).

\bibitem{zhang2019trafficspeedatt}
C.~Zhang, J.~James, Y.~Liu, Spatial-temporal graph attention networks: A deep
  learning approach for traffic forecasting, IEEE Access 7 (2019)
  166246--166256.

\bibitem{zhang2019trafficspeedgan}
Y.~Zhang, S.~Wang, B.~Chen, J.~Cao, Z.~Huang, Trafficgan: Network-scale deep
  traffic prediction with generative adversarial nets, IEEE Transactions on
  Intelligent Transportation Systems (2019).

\bibitem{yi2019trafficspeed}
X.~Yi, Z.~Duan, T.~Li, T.~Li, J.~Zhang, Y.~Zheng, Citytraffic: Modeling
  citywide traffic via neural memorization and generalization approach, in:
  Proceedings of the 28th ACM International Conference on Information and
  Knowledge Management, 2019, pp. 2665--2671.

\bibitem{cai2020trafficspeed}
L.~Cai, K.~Janowicz, G.~Mai, B.~Yan, R.~Zhu, Traffic transformer: Capturing the
  continuity and periodicity of time series for traffic forecasting,
  Transactions in GIS.

\bibitem{dataset:pems}
\url{http://pems.dot.ca.gov}.

\bibitem{dataset:qtraffic}
\url{https://ai.baidu.com/broad/download?dataset=traffic}.

\bibitem{dataset:taxinyc}
\url{https://www1.nyc.gov/site/tlc/about/tlc-trip-record-data.page}.

\bibitem{dataset:taxibj}
\url{https://github.com/TolicWang/DeepST/tree/master/data/TaxiBJ}.

\bibitem{dataset:bikenyc}
\url{https://www.citibikenyc.com/system-data}.

\bibitem{dataset:busnyc}
\url{http://web.mta.info/developers/MTA-Bus-Time-historical-data.html}.

\bibitem{dataset:webtris}
\url{https://data.gov.uk/dataset/9562c512-4a0b-45ee-b6ad-afc0f99b841f/ \\
  highways-england-network-journey-time-and-traffic-flow-data}.

\bibitem{dataset:drivenet}
\url{HTTP://wsdot.uwdrive.net/STARLab}.

\bibitem{dataset:porto}
\url{https://archive.ics.uci.edu/ml/datasets/Taxi+Service+Trajectory+-+ \\
  Prediction+Challenge,+ECML+PKDD+2015}.

\bibitem{dataset:didi}
\url{https://outreach.didichuxing.com/appEn-vue/personal?id=2}.

\bibitem{dataset:bikedc}
\url{https://www.capitalbikeshare.com/system-data}.

\bibitem{dataset:bikechi}
\url{https://www.divvybikes.com/system-data}.

\bibitem{dataset:hk}
\url{https://resource.data.one.gov.hk/td/speedmap.xml}.

\bibitem{dataset:uber}
\url{https://www.kaggle.com/fivethirtyeight/uber-pickups-in-new-york-city}.

\bibitem{dataset:topis}
\url{https://topis.seoul.go.kr/}.

\bibitem{dataset:buschi}
\url{https://data.cityofchicago.org/Transportation/Chicago-Traffic-Tracker-Historical-Congestion-Esti/77hq-huss/data}.

\bibitem{wolpert1997nolunch}
D.~H. Wolpert, W.~G. Macready, No free lunch theorems for optimization, IEEE
  transactions on evolutionary computation 1~(1) (1997) 67--82.

\bibitem{noruzoliaee2018roads}
M.~Noruzoliaee, B.~Zou, Y.~Liu, Roads in transition: Integrated modeling of a
  manufacturer-traveler-infrastructure system in a mixed autonomous/human
  driving environment, Transportation Research Part C: Emerging Technologies 90
  (2018) 307--333.

\end{thebibliography}

\end{document}